\documentclass[11pt]{article}

\usepackage[preprint]{acl}

\usepackage{times}
\usepackage{latexsym}
\usepackage[T1]{fontenc}
\usepackage[utf8]{inputenc}
\usepackage{microtype}
\usepackage{inconsolata}

\usepackage{graphicx}
\usepackage{booktabs}
\usepackage{amsfonts}
\usepackage{amsmath}
\usepackage{nicefrac}
\usepackage{xcolor}
\usepackage{enumitem}
\usepackage{fvextra}

\usepackage{tabularx}
\usepackage{array}

\title{HIVE: Hidden-Evidence Verification for Hallucination Detection in Diffusion Large Language Models
}

\author{
Guoshenghui Zhao$^{1}$ \quad Tan Yu$^{2}$ \quad Weijie Zhao$^{1}$\\
$^{1}$Rochester Institute of Technology \quad
$^{2}$NVIDIA Corporation\\
\texttt{\small gz1626@rit.edu \quad tayu@nvidia.com \quad wjz@cs.rit.edu}
}

\begin{document}

\maketitle

\begin{abstract}
Diffusion large language models generate text through iterative denoising, exposing hidden trajectories that may contain reliability signals beyond the final output. We propose HIVE, which compresses trajectory hidden states, selects informative step-layer evidence, and conditions a verifier through continuous prefix embeddings to produce a hallucination score and structured diagnostics. Across two D-LLMs and three QA benchmarks, HIVE outperforms eight established baselines and a verifier-backbone-matched text-only control in all six settings. Relative to text-only verification, hidden-evidence conditioning improves AUROC by 1.73--4.60 points and AUPRC by 1.10--3.62 points, with average gains of 3.15 and 2.28 points, respectively. Ablations, evidence interventions, and cross-dataset transfer further support the complementary value of fine-grained hidden trajectory evidence.
\end{abstract}

\section{Introduction}

Diffusion large language models (D-LLMs) generate text through iterative denoising rather than left-to-right next-token prediction \citep{li2022diffusionlm,gong2023diffuseq,lovelace2023latentdiffusion,lovelace2024diffusionguided,nie2025llada,ye2025dream7b}. This process exposes intermediate trajectory states in which token identities and internal representations evolve before the final answer is formed, creating reliability signals unavailable to output-only detectors.

\begin{figure}[t]
    \centering
    \includegraphics[width=\columnwidth]{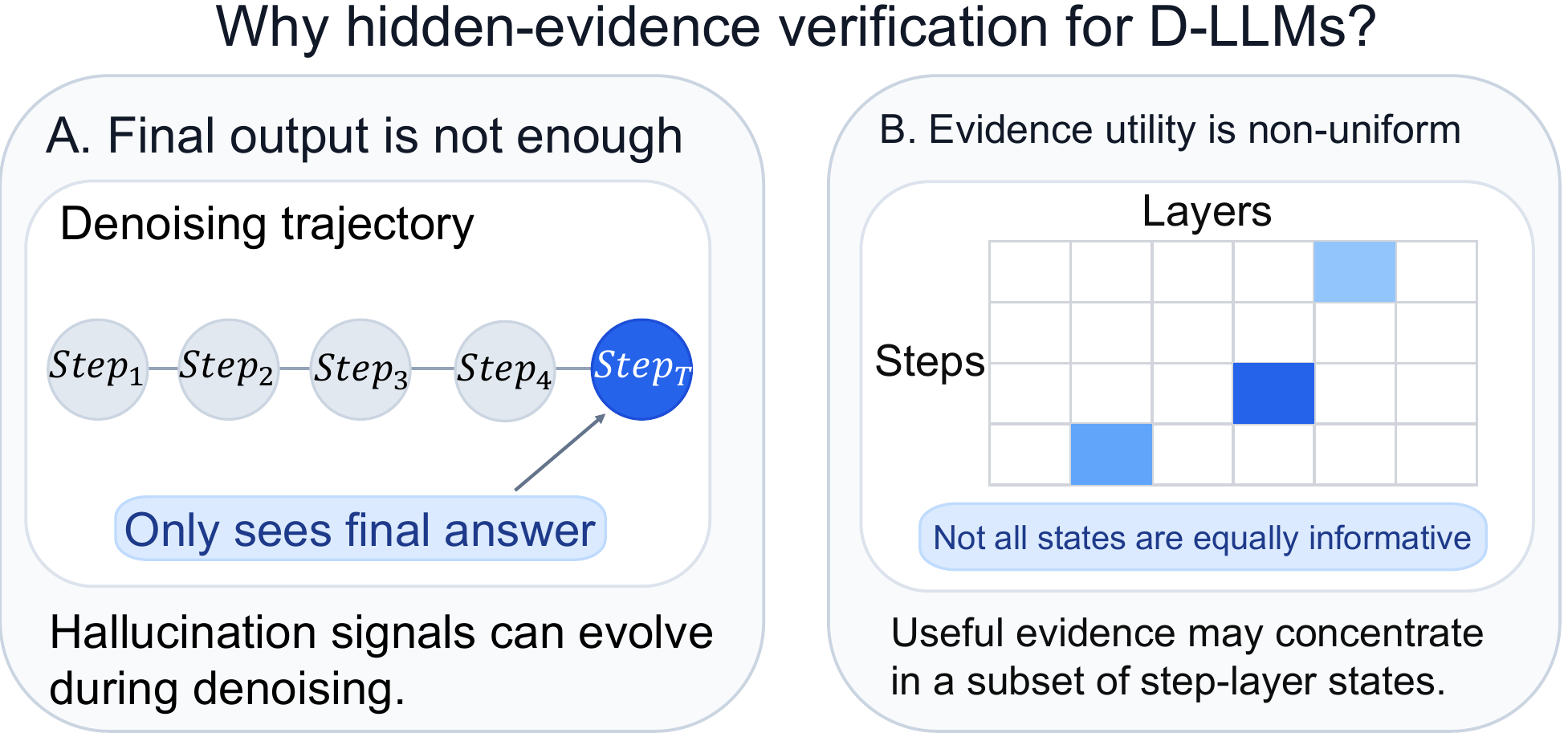}
    \caption{\textbf{Motivation for HIVE.} Hallucination-related signals evolve unevenly across the denoising trajectory; HIVE learns to select informative hidden step-layer states.}
    \label{fig:intro_teaser}
\end{figure}

Most hallucination detectors rely on output uncertainty, repeated-generation consistency, or static hidden representations \citep{zhang2025sirens,kadavath2022mostly,farquhar2024semanticentropy,kossen2024semanticentropyprobes,manakul2023selfcheckgpt,zhang2023sac3,zhao2024knowing,su2024mind,chen2024inside,zhang2025prism}. Recent studies show that hallucination-related signals also evolve across D-LLM denoising trajectories, but existing approaches mainly model sub-traces, token-level uncertainty, or attention dynamics \citep{chang2026tracedet,hemmat2026tdgnet,qian2026dynhd}. As illustrated in Figure~\ref{fig:intro_teaser}, they do not explicitly identify fine-grained informative hidden evidence jointly across denoising steps and model layers.

We propose HIVE, a hidden-evidence verification framework that compresses trajectory hidden states, selects informative step-layer evidence, and conditions a verifier through continuous prefix embeddings.\footnote{Code is available at \url{https://anonymous.4open.science/r/hive-dllm-D1C3/}.} HIVE produces both a continuous hallucination score and auxiliary structured diagnostics. We study answer-level hallucination detection in short-form QA, treating generated answers judged incorrect or unverifiable with respect to the question and available gold answers as hallucinated for evaluation.

Our main contributions are:
\begin{itemize}[leftmargin=*, itemsep=1pt, topsep=2pt]
    \item We introduce a fine-grained hidden-evidence perspective for hallucination detection in D-LLMs, focusing on informative states across denoising steps and layers rather than only final outputs or coarse trajectory summaries.

    \item We propose \textbf{HIVE}, which combines compressed two-stream hidden evidence, a learned step-layer selector, and an evidence-conditioned verifier. The verifier produces both a continuous hallucination score from decision logits and auxiliary structured diagnostic outputs.

    \item Across two D-LLMs and three QA benchmarks, HIVE outperforms eight established baselines and a verifier-backbone-matched text-only control in all six settings, with average gains of 3.15 AUROC and 2.28 AUPRC points. Cross-dataset transfer, evidence interventions, and component ablations further support the complementary value of selected hidden trajectory evidence.
\end{itemize}

\section{Related Work}

\paragraph{Hallucination detection.}
Existing hallucination detectors broadly rely on three types of signals: output uncertainty and calibration \citep{kadavath2022mostly,farquhar2024semanticentropy,kossen2024semanticentropyprobes}, consistency across repeated generations \citep{manakul2023selfcheckgpt,zhang2023sac3,zhao2024knowing}, and internal representations captured by learned probes or latent geometry \citep{su2024mind,chen2024inside,zhang2025prism}. Related work further studies semantic dispersion, factuality evaluation, and attention-based reliability signals \citep{lin2024confidence,vashurin2025cocoa,fadeeva2023lmpolygraph,min2023factscore,chuang2024lookback,vazhentsev2025unconditional}. Most of these methods are developed for autoregressive generation and operate on final outputs, multiple sampled responses, or static hidden representations, rather than fine-grained hidden evidence evolving across denoising steps and model layers.

\paragraph{D-LLM trajectories.}
D-LLMs expose denoising trajectories in which token identities and hidden representations evolve before output formation \citep{rossi2026lengthaware,kim2026earlydecisions,guo2026lostindiffusion,vonrutte2025scaling}. Recent detectors model sub-trajectories, uncertainty dynamics, or attention graphs but do not explicitly select fine-grained hidden units jointly across denoising steps and layers for downstream verification \citep{chang2026tracedet,hemmat2026tdgnet,qian2026dynhd}. HIVE addresses this gap by selecting fine-grained step-layer evidence for verification.

\section{Method}
\subsection{Overview of HIVE}

\begin{figure*}[t]
    \centering
    \includegraphics[width=0.95\textwidth]{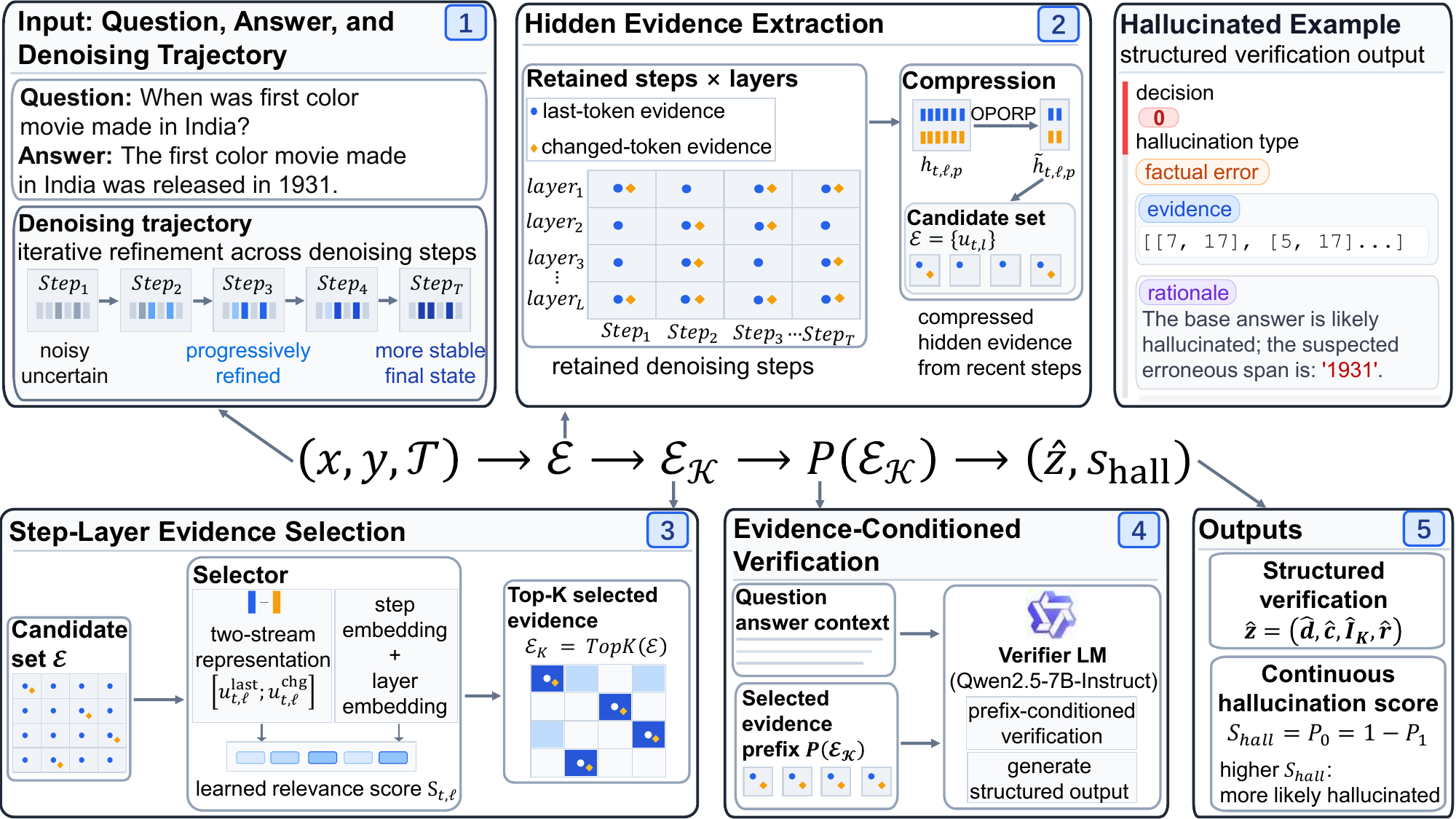}
    \caption{\textbf{Overview of HIVE.}
    Given a question, its D-LLM-generated answer, and the associated denoising trajectory, HIVE extracts compressed hidden evidence, selects informative step-layer evidence under a fixed budget, and conditions a verifier language model on the selected evidence through prefix embeddings. The verifier produces both a structured diagnostic output $\hat{z}$ and a continuous hallucination score $s_{\mathrm{hall}}$. This overall mapping is summarized in Eq.~\ref{eq:hive_pipeline}.}
    \label{fig:hive_overview}
\end{figure*}

Figure~\ref{fig:hive_overview} summarizes HIVE. Given a question $x$, a D-LLM-generated answer $y$, and its denoising trajectory $\mathcal{T}$, HIVE extracts candidate hidden evidence $\mathcal{E}$, selects a top-$K$ step-layer subset $\mathcal{E}_K$, and conditions a verifier through continuous prefix embeddings. The verifier produces a structured diagnostic output $\hat{z}$ and a continuous hallucination score $s_{\mathrm{hall}}$:
\begin{equation}
(x, y, \mathcal{T}) \;\longrightarrow\; \mathcal{E} \;\longrightarrow\; \mathcal{E}_K \;\longrightarrow\; (\hat{z}, s_{\mathrm{hall}}).
\label{eq:hive_pipeline}
\end{equation}
The following subsections describe evidence extraction, selection, verification, and decision-logit scoring.

\subsection{Hidden Evidence Extraction}

Let $\mathcal{T}$ denote the denoising trajectory associated with $(x,y)$, and let $h_{t,\ell,p} \in \mathbb{R}^{d}$ denote the hidden state at denoising step $t$, layer $\ell$, and token position $p$.

For each retained denoising step $t$, we define a candidate position set
$\mathcal{P}_t = \{p_t^{\mathrm{last}}\} \cup \mathcal{P}_t^{\mathrm{chg}}$, where $p_t^{\mathrm{last}}$ is the last token position of the current answer sequence and $\mathcal{P}_t^{\mathrm{chg}}$ contains positions whose decoded token identities differ from those at the previous denoising step. We retain only the last few denoising steps and cap the number of changed positions per step.

For each retained step $t$ and layer $\ell$, hidden states at positions in $\mathcal{P}_t$ are compressed into a lower-dimensional evidence space:
$\tilde{h}_{t,\ell,p} = \Pi_\ell(h_{t,\ell,p}) \in \mathbb{R}^{r}$, where $\Pi_\ell(\cdot)$ maps the original hidden state from dimension $d$ to a compressed evidence dimension $r \ll d$. In practice, we use OPORP random projection \citep{li2023oporp}.

For each retained step-layer pair $(t,\ell)$, let $H_{t,\ell}=\{\tilde{h}_{t,\ell,p}\mid p\in\mathcal{P}_t\}$ denote its compressed position-level hidden-state group. The next subsection maps each variable-sized group to a fixed-dimensional two-stream evidence vector used by both the selector and verifier.

\subsection{Step-Layer Evidence Selection}
\label{sec:step_layer_selection}

For each compressed position-level group $H_{t,\ell}$, we construct a fixed-dimensional evidence unit using two complementary streams. Specifically, we define $u_{t,\ell}^{\mathrm{last}}=\tilde{h}_{t,\ell,p_t^{\mathrm{last}}}\in\mathbb{R}^{r}$ and let $u_{t,\ell}^{\mathrm{chg}}$ be the mean of $\tilde{h}_{t,\ell,p}$ over $p\in\mathcal{P}_t^{\mathrm{chg}}$ when $|\mathcal{P}_t^{\mathrm{chg}}|>0$, and $\mathbf{0}$ otherwise.
The resulting evidence unit is $u_{t,\ell}=[u_{t,\ell}^{\mathrm{last}};u_{t,\ell}^{\mathrm{chg}}]\in\mathbb{R}^{2r}$.
Let $\mathcal{S}$ denote the retained denoising-step index set and $\mathcal{R}$ denote the retained model-layer index set. We collect all retained evidence units into
\[
\mathcal{E}
=
\left\{
u_{t,\ell}
\mid
t\in\mathcal{S},\,
\ell\in\mathcal{R}
\right\}.
\]
Since $\mathcal{E}$ contains one candidate for every retained step-layer pair, HIVE learns a selector that jointly scores these units and retains the most informative top-$K$ subset under a fixed evidence budget.

To account for the different roles that similar hidden patterns may play across denoising steps and model layers, the selector explicitly incorporates step-layer identity. Let $a_t\in\mathbb{R}^{d_{\mathrm{step}}}$ and $b_\ell\in\mathbb{R}^{d_{\mathrm{layer}}}$ denote learnable step and layer embeddings, respectively. For each candidate pair $(t,\ell)$, we concatenate the two-stream evidence feature with these embeddings:
\[
\tilde{u}_{t,\ell}
=
\left[
u_{t,\ell};
a_t;
b_\ell
\right]
\in
\mathbb{R}^{2r+d_{\mathrm{step}}+d_{\mathrm{layer}}}.
\]
A shared multilayer perceptron then assigns a scalar relevance score to each candidate unit:
\[
s_{t,\ell}
=
f_{\mathrm{gate}}\!\left(\tilde{u}_{t,\ell}\right).
\]

Let $\mathcal{J}=\{(t,\ell)\mid t\in\mathcal{S},\,\ell\in\mathcal{R}\}$ denote the index set of retained step-layer units. We normalize the relevance scores over all candidate units using temperature $\tau>0$:
\[
\pi_{t,\ell}
=
\frac{\exp(s_{t,\ell}/\tau)}
{\sum_{(t',\ell')\in\mathcal{J}}\exp(s_{t',\ell'}/\tau)},
\qquad
(t,\ell)\in\mathcal{J}.
\]
During training, HIVE applies a straight-through top-$K$ operator that retains the $K$ highest-scoring pairs in the forward pass while propagating gradients through the corresponding soft score distribution. Let $\tilde{w}_{t,\ell}$ denote the resulting straight-through selection weights. The aggregated selector representation is
\[
m
=
\sum_{(t,\ell)\in\mathcal{J}}
\tilde{w}_{t,\ell}\tilde{u}_{t,\ell}.
\]
A lightweight hallucination probe predicts the example-level binary label from $m$, providing supervision for learning the selector scores. An entropy regularization term is applied to the selected weights to stabilize the learned evidence distribution.

After training, the selector is frozen and the top-$K$ units are exported for verifier training. Let $\mathcal{I}_K=((t_j,\ell_j))_{j=1}^{K}$ denote the ordered list of the $K$ highest-scoring step-layer pairs:
\[
\mathcal{I}_K
=
\operatorname*{arg\,topK}_{(t,\ell)\in\mathcal{J}}
s_{t,\ell},
\qquad
\mathcal{E}_K
=
\left(u_{t_j,\ell_j}\right)_{j=1}^{K},
\]
where the selected pairs are ordered by decreasing relevance score. For each selected pair $(t_j,\ell_j)\in\mathcal{I}_K$, we renormalize the selector probabilities over the retained top-$K$ set:
\[
\omega_j
=
\frac{\pi_{t_j,\ell_j}}
{\sum_{(t',\ell')\in\mathcal{I}_K}\pi_{t',\ell'}},
\qquad
(t_j,\ell_j)\in\mathcal{I}_K.
\]
Thus, $\mathcal{E}_K$ contains the $K$ highest-scoring evidence units selected jointly across the retained denoising steps and model layers, and $\omega_j$ denotes the normalized selector weight exported with the $j$-th selected unit.

\subsection{Evidence-Conditioned Verification}

HIVE conditions a verifier language model on both the original question--answer context and the selected hidden evidence. For each selected pair $(t_j,\ell_j)\in\mathcal{I}_K$, the verifier-side evidence input is the corresponding two-stream representation $u_{t_j,\ell_j}\in\mathbb{R}^{2r}$, together with its normalized selector weight, step identity, and layer identity. Let $a^{\mathrm{ver}}_{t_j}$ and $b^{\mathrm{ver}}_{\ell_j}$ denote the verifier-side step and layer embeddings for the $j$-th selected pair. The evidence encoder maps these aligned signals into the verifier hidden space:
\[
v_j
=
\psi_{\mathrm{ver}}
\left(
u_{t_j,\ell_j},
\omega_j,
a^{\mathrm{ver}}_{t_j},
b^{\mathrm{ver}}_{\ell_j}
\right)
\in
\mathbb{R}^{d_{\mathrm{ver}}}.
\]
Collecting the encoded evidence units yields the hidden-evidence prefix
\[
P(\mathcal{E}_K)
=
\left[
v_1,\ldots,v_K
\right].
\]

The resulting sequence is prepended to the textual token embeddings as a continuous hidden-evidence prefix, while the textual input remains the question--answer context $(x,y)$. Conditioned on $(x,y)$ and $P(\mathcal{E}_K)$, the verifier generates a structured diagnostic output
\[
\hat{z} = \mathrm{Ver}\bigl(x, y \,;\, P(\mathcal{E}_K)\bigr),
\]
which contains a binary hallucination decision, a predicted hallucination type, references to the selected step-layer units, and a short rationale. Concretely, we write
$\hat{z} = \bigl(\hat{d}, \hat{c}, \hat{\mathcal{I}}_K, \hat{r}\bigr)$,
where $\hat{d} \in \{0,1\}$ is the verifier decision, $\hat{c}$ is the predicted hallucination type,  $\hat{\mathcal{I}}_K$ denotes the reproduced references to the selected step-layer units, and $\hat{r}$ is a short
diagnostic rationale. During training, the verifier is supervised to generate this structured output format.

\subsection{Decision-Logit Hallucination Scoring}
\label{sec:decision_logit_scoring}

Let $z_0$ and $z_1$ denote the verifier logits for decision labels 0 (hallucinated) and 1 (non-hallucinated) at the decision-emission step. We define
\[
s_{\mathrm{hall}}
=
\frac{\exp(z_0)}{\exp(z_0)+\exp(z_1)}.
\]
Higher values indicate greater hallucination likelihood. Unlike the hard decision, this continuous score preserves ranking information and supports AUROC and AUPRC.

\subsection{Training Procedure}
\label{sec:training_procedure}

HIVE is trained in two stages. First, the selector is trained jointly with an auxiliary hallucination probe using example-level labels, then frozen to export top-$K$ evidence. Second, the evidence encoder and verifier are trained jointly on the exported evidence and question--answer text.

\section{Experimental Setup}

\subsection{Models and Benchmarks}

We evaluate HIVE in six separately trained model--benchmark settings using two representative D-LLMs: Dream-v0-Instruct-7B, referred to as Dream-7B-Instruct, and LLaDA-8B-Instruct.

We consider three QA benchmarks with complementary characteristics: TriviaQA, HotpotQA, and NQOpenLike \citep{joshi2017triviaqa,yang2018hotpotqa,kwiatkowski2019naturalquestions}. TriviaQA provides an open-domain QA setting, HotpotQA emphasizes multi-hop reasoning, and NQOpenLike is an NQ-derived open-ended QA collection.

\subsection{Data Construction and Supervision}
\label{sec:data_supervision}

Our evaluation target covers factual errors, fabrications, unsupported claims, incomplete or irrelevant responses, and reasoning errors in short-form QA, rather than only fabricated factual content. For each benchmark, we use the original questions and all available gold-answer variants. Each target D-LLM generates one base answer per question, while its associated denoising trajectory is recorded for hidden-evidence extraction. Evaluating all three benchmarks with both D-LLMs yields six model--benchmark settings, each with a separately trained HIVE system and a predefined question-level split. Dataset sizes, splits, and label distributions are reported in Appendix Table~\ref{tab:data_statistics}.

The selector and verifier use the same post-processed binary correctness target, initialized from an LLM judge and potentially revised by deterministic consistency checks against the available gold answers. We use Qwen2.5-7B-Instruct as the judge with greedy decoding and a maximum generation length of 160 tokens. The judge receives the question, all available gold answers, and the D-LLM-generated answer. It assigns an initial correctness label of 1 only when the generated answer is judged semantically equivalent to at least one valid gold answer; answers that are incorrect or whose correctness cannot be established receive a label of 0. The judge additionally predicts a confidence value, hallucination type, minimal verbatim suspect span, and short rationale.

The parsed judge output is normalized and subjected to deterministic gold-answer consistency checks. Correct examples are assigned type \emph{none} with an empty suspect span. Questions requiring multiple answer elements are checked for sufficient answer coverage, and strict or soft gold-answer evidence may revise raw judge decisions when they conflict with deterministic matching evidence. Invalid JSON outputs conservatively receive an incorrect label, with type \emph{incomplete} when gold answers are available and \emph{unsupported} otherwise. We refer to the resulting target throughout the paper as the post-processed judge label.

\subsection{Baselines}

We compare HIVE against eight representative baselines spanning output-based, latent-based, and trajectory-based detection, together with a verifier-backbone-matched text-only verifier control.

\textbf{Output-based baselines.} We include Perplexity, LN-Entropy, Semantic Entropy, and Lexical Similarity. These methods estimate hallucination risk from the generated outputs themselves, without explicitly using intermediate hidden trajectories. Perplexity and LN-Entropy rely on token-level or sequence-level uncertainty signals \citep{malinin2021uncertainty}. Semantic Entropy and Lexical Similarity, by contrast, measure the consistency of multiple sampled answers at the semantic or lexical level \citep{kuhn2023semanticuncertainty}.

\textbf{Latent-based baselines.} We include EigenScore, CCS, and TSV, which operate on internal model representations rather than output-only uncertainty \citep{chen2024inside,burns2023latent,park2025steer}. These methods test whether hidden-state geometry or learned latent probes can provide stronger hallucination signals than output-level statistics alone \citep{azaria2023internal,du2024haloscope}.

\textbf{Trajectory-based baseline.}
We include TraceDet, a recent diffusion-language-model hallucination detector that operates on decoding trajectories \citep{chang2026tracedet}. TraceDet is the most directly related established baseline in our comparison, as it also exploits trajectory-level signals in D-LLMs. Compared with TraceDet, HIVE explicitly selects fine-grained hidden evidence across denoising steps and model layers and conditions a verifier on the selected evidence.

\textbf{Verifier-backbone-matched text-only control.}
We additionally train a text-only verifier separately for each model--benchmark setting. This control uses the same Qwen2.5-7B-Instruct verifier backbone, primary correctness supervision, question-level splits, LoRA configuration, verifier-side optimization schedule, checkpoint-selection procedure, and decision-logit scoring protocol as HIVE. It receives only the question--answer text and omits the hidden-evidence encoder, continuous evidence prefix, and step-layer inputs. This comparison isolates the contribution of selected trajectory evidence from the effects of verifier-backbone capacity and supervised verifier fine-tuning.

\textbf{Common training and calibration protocol.}
All trainable baselines and verifier controls use the same predefined question-level train, validation, and test assignments. Model parameters and checkpoints are learned or selected without access to the test split. Decision thresholds are selected on the corresponding validation split by maximizing F1 and are transferred unchanged to the test split. Detailed score definitions, sampling configurations, representation pooling rules, and training settings are provided in Appendix~\ref{app:baseline_impl}.

\subsection{Evaluation Metrics}

We report AUROC and AUPRC as the primary threshold-free detection metrics \citep{davis2006prroc}. For both HIVE and the verifier-backbone-matched text-only control, the continuous hallucination score is derived from the verifier's decision-emission logits following Section~\ref{sec:decision_logit_scoring}. We additionally report F1, Balanced Accuracy, and Specificity using thresholds selected on validation data and transferred unchanged to test data \citep{brodersen2010balanced}. For HIVE's structured outputs, we report valid-JSON rate, decision accuracy, hallucination-type accuracy, and full-schema exact match. Hallucinated answers are treated as the positive class. Because NQOpenLike is more label-imbalanced, we emphasize AUROC and AUPRC in that setting. Additional thresholded and structured-output results are provided in Appendix~\ref{app:complete_results}.

\subsection{Implementation Details}

Following the staged procedure in Section~\ref{sec:training_procedure}, HIVE first trains and freezes the step-layer selector and then trains the evidence-conditioned verifier on the exported top-$K$ evidence. Dream-7B-Instruct generates answers using 64 denoising steps, whereas LLaDA-8B-Instruct uses 32 steps. Both models generate at most 32 answer tokens with temperature $0.0$ and top-$p$ $0.95$. For trajectory construction, we retain the final eight forward calls, capture the last-token position together with at most 16 changed-token positions per retained forward, and collect hidden states from all 28 Dream layers or all 32 LLaDA layers. We compress the captured hidden states using OPORP random projection to dimension $r=64$.

For step-layer evidence selection, we concatenate each two-stream feature with separate 16-dimensional learnable step and layer embeddings and score the resulting representation with a shared multilayer perceptron, as described in Section~\ref{sec:step_layer_selection}. We select a fixed evidence budget of $K=16$ step-layer units per example. For evidence-conditioned verification, we use Qwen2.5-7B-Instruct as the verifier backbone \citep{qwen2025technical}; the selected evidence units are encoded into continuous prefix embeddings and prepended to the original question--answer context, following continuous prefix conditioning \citep{li2021prefixtuning}. The verifier is trained to generate a structured output containing a binary hallucination decision, a predicted hallucination type, references to the selected step-layer units, and a short rationale. HIVE derives its continuous hallucination score from the verifier's decision-emission logits, with thresholds selected on the validation split and transferred unchanged to the test split. Additional implementation details, prompt templates, runtime hyperparameters, dataset statistics, baseline and control configurations, and inference profiling are provided in Appendices~\ref{app:exp_details},~\ref{app:prompt_schema},~\ref{app:baseline_impl},~\ref{app:additional_results}, and~\ref{app:inference_profiling}.

\section{Main Results}

\subsection{In-Domain Detection Results}
\label{sec:in_domain_results}

\begin{table*}[t]
\centering
\small
\setlength{\tabcolsep}{4pt}
\resizebox{\textwidth}{!}{
\begin{tabular}{lcccccc}
\toprule
& \multicolumn{3}{c}{\textbf{Dream-7B-Instruct}}
& \multicolumn{3}{c}{\textbf{LLaDA-8B-Instruct}} \\
\cmidrule(lr){2-4}\cmidrule(lr){5-7}
\textbf{Method}
& \textbf{HotpotQA}
& \textbf{NQOpenLike}
& \textbf{TriviaQA}
& \textbf{HotpotQA}
& \textbf{NQOpenLike}
& \textbf{TriviaQA} \\
\midrule
Semantic Entropy
& 81.84 / 93.10
& 72.17 / 91.47
& 81.92 / 86.92
& 78.00 / 90.04
& 81.24 / 94.57
& 80.98 / 86.79 \\

CCS
& 89.60 / 96.65
& 81.31 / 92.98
& 88.75 / 91.22
& 87.79 / 92.98
& 81.46 / 95.09
& 84.33 / 89.95 \\

TraceDet
& 87.40 / 95.64
& 79.47 / 94.86
& 83.44 / 88.60
& 86.05 / 94.28
& 77.68 / 94.16
& 80.07 / 86.89 \\
\midrule
Text-only verifier
& 89.42 / 96.27
& 83.61 / 95.39
& 88.54 / 91.75
& 89.21 / 94.73
& 81.11 / 94.05
& 86.40 / 90.90 \\

\textbf{HIVE}
& \textbf{91.76 / 97.37}
& \textbf{85.74 / 96.60}
& \textbf{92.36 / 95.37}
& \textbf{90.94 / 96.48}
& \textbf{85.71 / 96.95}
& \textbf{90.66 / 93.98} \\
\bottomrule
\end{tabular}}
\caption{\textbf{Compact in-domain detection results.} Each cell reports AUROC / AUPRC in percent. For compactness, we show the strongest output-based and latent-based baselines, the trajectory-based TraceDet baseline, and the verifier-backbone-matched text-only control. Complete results for all eight baselines are provided in Appendix Table~\ref{tab:appendix_full_main_results}.}
\label{tab:main_results_compact}
\end{table*}

\begin{figure*}[t!]
\centering
\includegraphics[width=0.97\textwidth]{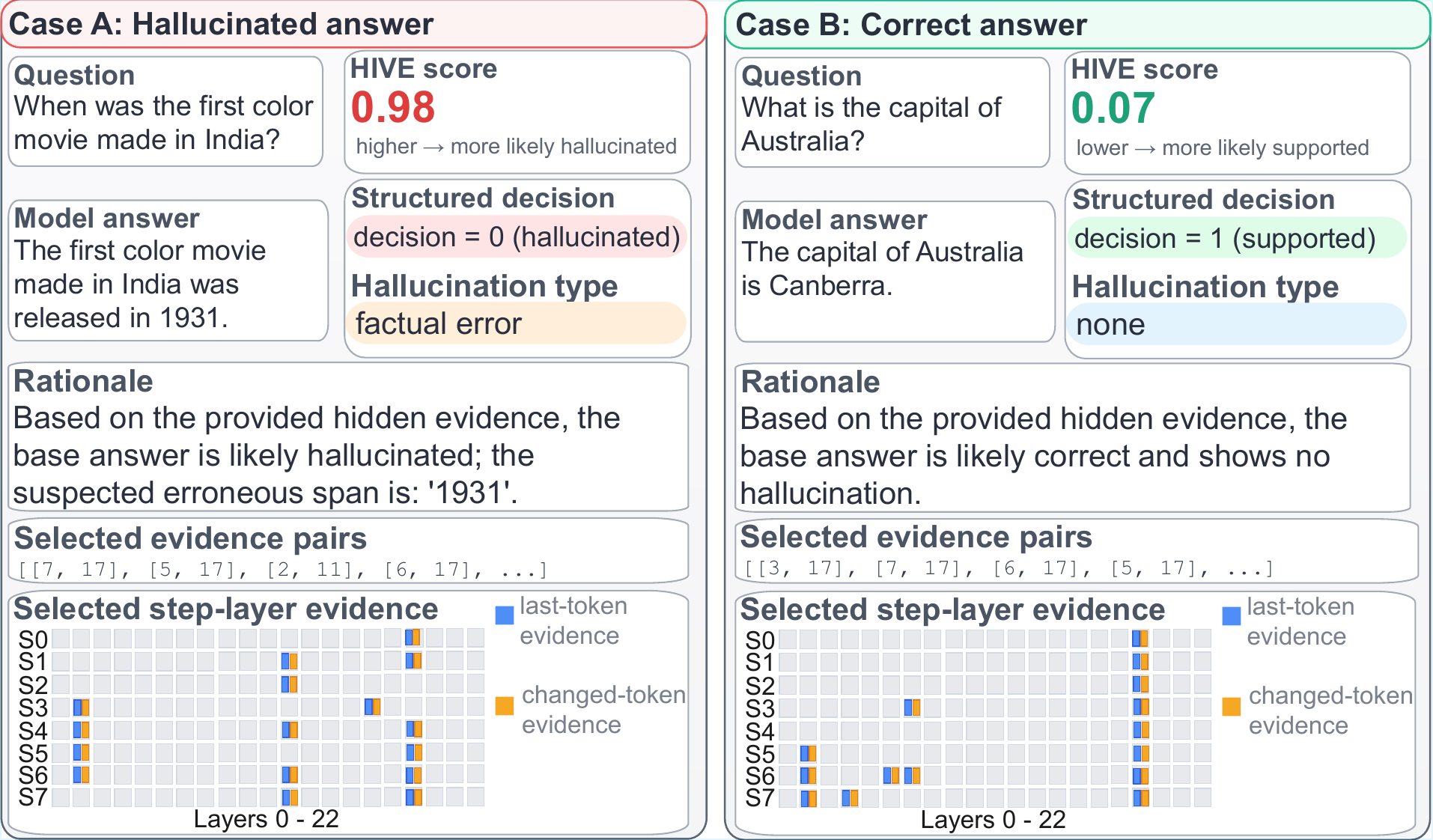}
\caption{\textbf{Representative qualitative outputs of HIVE.} Hallucinated and non-hallucinated examples with the HIVE score, structured decision, predicted hallucination type, diagnostic rationale, selected step-layer references, and corresponding selection pattern.}
\label{fig:qual_cases}
\end{figure*}

Table~\ref{tab:main_results_compact} shows that HIVE achieves the best AUROC and AUPRC across all six in-domain settings. Relative to the verifier-backbone-matched text-only verifier, HIVE improves AUROC by 1.73--4.60 points and AUPRC by 1.10--3.62 points, with average gains of 3.15 and 2.28 points, respectively. The improvements hold across both D-LLM families and all three benchmarks, indicating that selected hidden trajectory evidence contributes information beyond the question--answer text.

Figure~\ref{fig:qual_cases} illustrates representative hallucinated and non-hallucinated outputs, including their continuous scores, structured predictions, and selected step-layer patterns. Additional thresholded classification and structured-output results are reported in Appendix Tables~\ref{tab:appendix_full_thresholded_results} and~\ref{tab:appendix_full_structured_results}.

\subsection{Input-Source Decomposition}
\label{sec:component_comparison}

To disentangle the contributions of textual and hidden inputs, we perform an input-source decomposition on Dream-7B-Instruct + TriviaQA. We compare zero-shot textual verification, a hidden-only selector probe, a separately fine-tuned text-only verifier, and Full HIVE. Full HIVE exceeds the text-only verifier by 3.82 AUROC and 3.62 AUPRC points and the hidden-only selector probe by 7.59 AUROC and 5.53 AUPRC points. These results show that textual verification and selected hidden trajectory evidence provide complementary rather than interchangeable signals. Complete results, including F1, are reported in Appendix Table~\ref{tab:appendix_component_comparison}.

\subsection{Cross-Dataset Transfer}
\label{sec:cross_dataset_transfer}

We evaluate cross-dataset transfer within short-form QA using Dream-7B-Instruct as the fixed target D-LLM. The selector, hidden-evidence prefix encoder, and verifier are trained only on the TriviaQA training split, with checkpoint selection performed only on the corresponding validation split. The source-trained text-only verifier follows the same verifier-side training and scoring protocol but receives only the question--answer text, without trajectory-derived evidence, step-layer identifiers, or continuous prefix inputs. Both systems are then applied directly to the HotpotQA and NQOpenLike test sets without target-side training, validation, threshold calibration, or checkpoint selection.

\begin{table*}[t]
\centering
\small
\setlength{\tabcolsep}{6pt}
\renewcommand{\arraystretch}{1.05}
\begin{tabular}{@{}lccc@{}}
\toprule
\textbf{Method} & \textbf{TriviaQA (ID)} & \textbf{HotpotQA (OOD)} & \textbf{NQOpenLike (OOD)} \\
\midrule
Trivia-trained text-only verifier & 88.54 / 91.75 & 82.47 / 93.31 & 75.68 / 92.39 \\
\textbf{Trivia-trained HIVE} & \textbf{92.36 / 95.37} & \textbf{85.32 / 94.65} & \textbf{80.78 / 94.53} \\
\midrule
Target-trained text-only verifier (reference)
& 88.54 / 91.75
& 89.42 / 96.27
& 83.61 / 95.39 \\

Target-trained HIVE (reference)
& 92.36 / 95.37
& 91.76 / 97.37
& 85.74 / 96.60 \\
\bottomrule
\end{tabular}
\caption{\textbf{Cross-dataset transfer within short-form QA using Dream-7B-Instruct.} Each cell reports AUROC / AUPRC in percent. Bold indicates the stronger source-trained method. Target-trained text-only and HIVE systems are included only as in-domain references and are not part of the transfer comparison.}
\label{tab:cross_dataset_transfer}
\end{table*}

Relative to the Trivia-trained text-only verifier, source-trained HIVE improves AUROC by 3.82 points on TriviaQA, 2.85 on HotpotQA, and 5.10 on NQOpenLike, with AUPRC gains of 3.62, 1.34, and 2.14 points, respectively. On the two OOD benchmarks, both source-trained systems remain below their corresponding target-trained references, indicating that dataset-specific training remains beneficial. Nevertheless, HIVE consistently outperforms the matched source-trained text-only verifier, showing that selected hidden evidence remains complementary under dataset shift.

\subsection{Label Reliability}
\label{sec:label_reliability}

On a balanced 120-example audit spanning all six model--benchmark settings, the blinded two-annotator evaluation supports the reliability of the primary binary supervision: annotator agreement is 94.17\%, and the automatic labels achieve 92.50\% accuracy and 92.46 macro-F1 against the adjudicated human labels. Hallucination-type annotations show lower agreement and lower auto--human alignment and are therefore treated only as auxiliary diagnostic targets. The complete audit table and protocol are reported in Appendix Table~\ref{tab:human_label_audit} and Appendix~\ref{app:human_audit}.

\subsection{Inference Efficiency}

Full HIVE incurs a moderate deployment cost, increasing mean latency by 31.48\% and peak GPU memory by 11.50\% relative to the verifier-backbone-matched text-only pipeline. Complete end-to-end and component-level profiling is provided in Appendix~\ref{app:inference_profiling}.

\section{Evidence Intervention and Ablation}

\subsection{Second-Setting Removal and Replacement Test}
\label{sec:second_setting_interventions}

To test whether dependence on selected hidden evidence extends beyond the primary ablation setting, we intervene on a trained LLaDA-8B-Instruct + HotpotQA system. \emph{No prefix} removes all hidden-evidence inputs at evaluation time, while \emph{Random-$K$} replaces the learned top-$K$ evidence with $K$ randomly selected step-layer units under the same budget. The question--answer text and verifier checkpoint remain unchanged.

\begin{table}[t]
\centering
\small
\setlength{\tabcolsep}{7pt}
\renewcommand{\arraystretch}{1.05}
\begin{tabular}{@{}lrr@{}}
\toprule
\textbf{Variant} & \textbf{AUROC} & \textbf{AUPRC} \\
\midrule
\textbf{Full HIVE} & \textbf{90.94} & \textbf{96.48} \\
No prefix at inference & 85.71 & 91.42 \\
Random-$K$ evidence & 84.86 & 90.73 \\
\bottomrule
\end{tabular}
\caption{\textbf{Evidence interventions on LLaDA-8B-Instruct + HotpotQA.} Results are percentages; higher is better.}
\label{tab:second_setting_ablation}
\end{table}

Removing the prefix reduces AUROC by 5.23 points and AUPRC by 5.06 points, while random replacement produces larger drops of 6.08 and 5.75 points. Together, these results show that HIVE benefits from aligned, selectively chosen hidden evidence rather than from prefix capacity alone.

\subsection{Primary-Setting Component Ablations}
\label{sec:primary_setting_ablations}

\begin{figure}[t]
\centering
\includegraphics[width=\columnwidth]{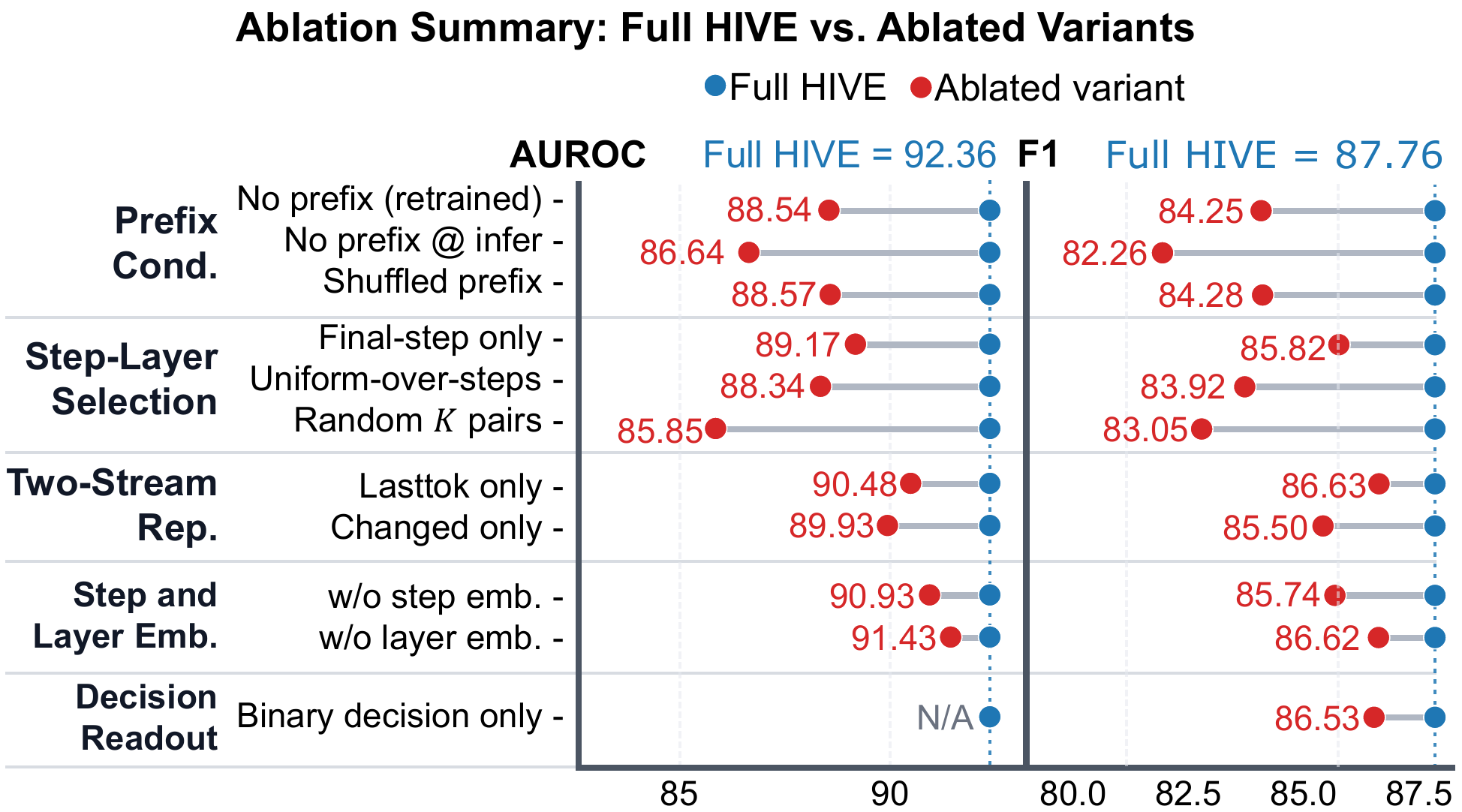}
\caption{\textbf{Ablation summary on Dream-7B-Instruct + TriviaQA.} Full HIVE is compared with representative variants under AUROC and F1. Complete per-variant results are reported in Appendix Table~\ref{tab:appendix_full_ablation_results}.}
\label{fig:ablation_summary}
\end{figure}

Figure~\ref{fig:ablation_summary} summarizes the primary-setting ablations. Breaking aligned prefix conditioning or restricting evidence selection produces the largest losses, while removing one evidence stream or explicit step-layer identity also causes consistent degradation. Complete AUROC, AUPRC, F1, balanced accuracy, specificity, and decision-accuracy results for all variants are reported in Appendix Table~\ref{tab:appendix_full_ablation_results}, with additional analysis in Appendix~\ref{app:detailed_ablations}.

\begin{figure}[t]
\centering
\includegraphics[width=\columnwidth]{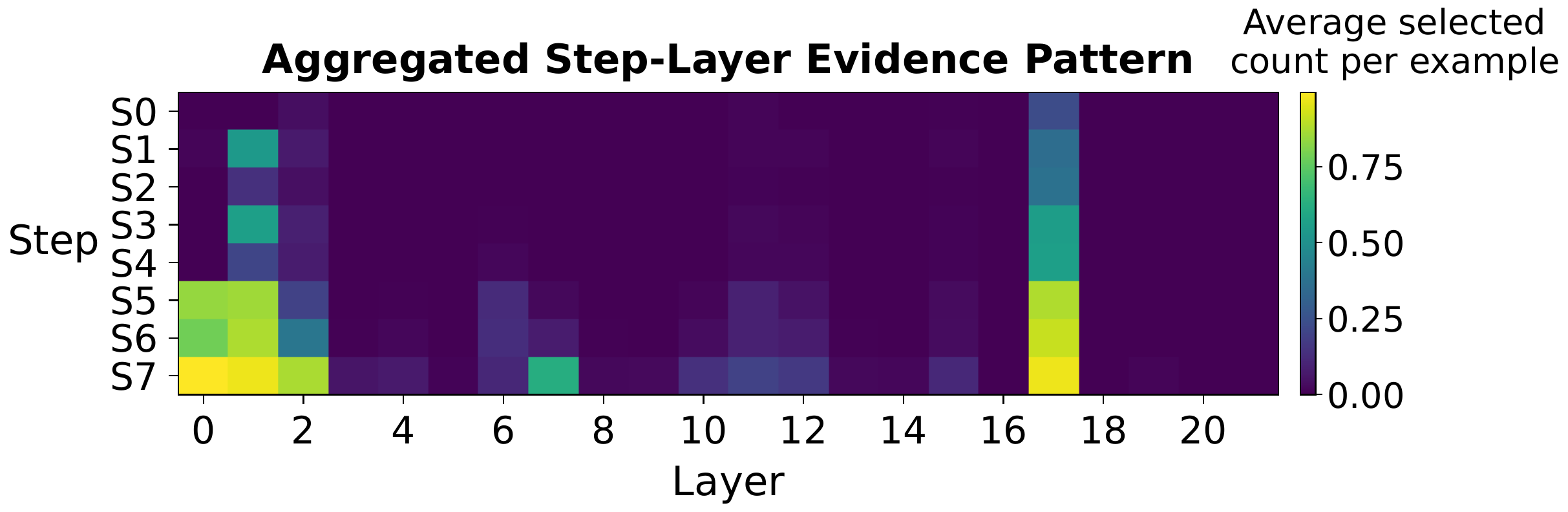}
\caption{\textbf{Aggregated step-layer evidence pattern.} Average selected count per example over the test set. The selector concentrates evidence on a sparse subset of denoising steps and layers rather than distributing it uniformly.}
\label{fig:agg_step_layer_pattern}
\end{figure}

The aggregate pattern in Figure~\ref{fig:agg_step_layer_pattern} complements the selection ablations: useful evidence is distributed across the trajectory but is concentrated in particular step-layer regions. The \emph{Final-step only}, \emph{Uniform over steps}, and \emph{Random-$K$ evidence} variants all underperform the learned selector, indicating that both trajectory coverage and learned prioritization matter.

\section{Conclusion}

We presented \textbf{HIVE}, a framework that selects hidden step-layer evidence from D-LLM trajectories and conditions a verifier through continuous prefix embeddings. Across two D-LLMs and three QA benchmarks, HIVE achieves the strongest performance among evaluated methods and outperforms a verifier-backbone-matched text-only control in all six settings by 3.15 AUROC and 2.28 AUPRC points on average. Evidence interventions, component ablations, and cross-dataset transfer show that selected hidden trajectory evidence provides complementary reliability signals beyond textual verification, output uncertainty, and coarse trajectory summaries.

\section{Limitations}

HIVE is evaluated on two D-LLM families and three short-form QA benchmarks, with transfer assessed only across datasets under a fixed Dream backbone. Broader evaluation across additional D-LLMs, tasks, and model families is needed to establish wider generality. HIVE also introduces additional inference cost, primarily from evidence-conditioned verifier decoding, and the reported detection results are based on single training runs. The selected step-layer references, generated rationales, and hallucination-type predictions should be viewed as auxiliary diagnostics rather than validated causal explanations, with type labels showing lower human alignment than binary correctness labels. Finally, using Qwen2.5-7B-Instruct for both automatic judging and verification may introduce shared-backbone bias, which future work should examine using independent judge and verifier architectures.

\section{Broader Impacts}

HIVE may support reliability assessment and structured diagnostic reporting for diffusion language models. However, its predictions remain fallible and should not be treated as guarantees of correctness, particularly in high-stakes applications. Human oversight and task-specific validation are therefore necessary before deployment.

\bibliography{references}

\appendix

\section{Additional Experimental Details}
\label{app:exp_details}

\subsection{End-to-End Pipeline Overview}
Our experiments follow a six-stage pipeline that instantiates HIVE from raw question-answer data to final hallucination detection metrics. At a high level, this pipeline refines the four-component formulation in the main text---hidden evidence extraction, step-layer evidence selection, evidence-conditioned verification, and decision-logit scoring---into a reproducible sequence of preprocessing, training, export, and evaluation stages.

In the first stage, we run the target diffusion language model on each input question and record its denoising trajectory. Rather than storing the full hidden trajectory, we retain only a fixed number of recent denoising forwards and extract hidden states from a sparse set of candidate positions at each retained step, consisting of the last-token position and a capped set of changed-token positions. These hidden states are then compressed with OPORP random projection and written to memory-mapped feature files together with auxiliary metadata such as row IDs, capture lengths, and forward IDs.

In the second stage, we train a step-layer-aware selector on top of the compressed trajectory features. For each candidate step-layer unit, we construct a two-stream representation that concatenates a last-token feature with a mean-pooled changed-token feature. The selector augments this two-stream evidence with learnable step and layer embeddings, scores all retained step-layer units jointly, and learns a top-$K$ evidence policy under hallucination supervision. In practice, the selector is trained with a straight-through top-$K$ mechanism and an entropy regularizer to encourage stable evidence selection.

In the third stage, we apply the trained selector to each split and export the selected evidence for downstream verification. For every example, we store the selected step-layer pairs, their top-$K$ selector weights, and the alignment information needed to reconstruct the evidence units. When requested, we also export the selected compressed activations themselves as \texttt{act\_top}, together with the derived two-stream evidence representation used by the verifier.

In the fourth stage, we pack the exported evidence into verifier training data. Each packed example contains the original question, the base model answer, the selected step-layer pairs, and a structured JSON target specifying the hallucination decision, hallucination type, evidence field, and rationale. This yields train/validation/test JSONL files for the verifier, while preserving pointers back to the exported evidence shards for efficient loading.

In the fifth stage, we train a verifier language model with hidden-evidence prefix conditioning. The verifier uses Qwen2.5-7B-Instruct as the textual backbone. Each selected evidence unit is converted into a two-stream evidence vector, encoded into the verifier hidden space, and injected as a prefix embedding sequence before the textual question-answer context. The verifier is then trained to generate a schema-constrained structured output, allowing HIVE to perform verification rather than only binary scoring.

In the final stage, we evaluate the trained verifier on held-out examples. During inference, the verifier generates the structured JSON prediction, and we additionally extract the decision-emission logits associated with the binary decision field. These logits are converted into a continuous hallucination score, which is used to compute threshold-free metrics such as AUROC and AUPRC, while the structured prediction itself is used to evaluate decision accuracy, Hallucination-Type Accuracy, valid-JSON rate, and other field-level exact-match metrics.

The remaining subsections provide the implementation details for each part of this pipeline, including feature storage, selector training, exported evidence format, verifier supervision, evaluation protocol, and runtime hyperparameters.

\subsection{Training, Validation, and Transfer Splits}
\label{app:split_protocol}

For each model--benchmark setting, we use a fixed question-level train, validation, and test split. The training split is used to learn the HIVE selector, auxiliary probe, evidence encoder, and verifier, and to train the separately initialized text-only verifier control. The validation split is used for checkpoint selection, early stopping, and decision-threshold calibration for all trainable systems, while the test split is held out from both parameter learning and calibration. These assignments are preserved throughout trajectory feature construction, selector training, evidence export, verifier training, and final evaluation.

For cross-dataset transfer, all trainable HIVE components and the text-only verifier baseline are trained only on the Dream-7B-Instruct + TriviaQA training split, with checkpoint selection performed only on the corresponding TriviaQA validation split. HotpotQA and NQOpenLike are used only for final test-set evaluation, without target-side training, validation, checkpoint selection, or threshold calibration.

\subsection{Data Construction and Trajectory Feature Storage}

We construct trajectory features before selector training by running the target D-LLM on each input question and recording a compact representation of its denoising process. In our implementation, the builder reads question-answer supervision from a tab-separated file in which each line contains a question and its gold answer list, split on the first tab only. For each example, the script formats the question into a chat-style prompt, runs the base diffusion language model to obtain a generated answer, and records the corresponding denoising trajectory together with the generated text and label metadata. The same builder supports both the native diffusion-generation interface and a local LLaDA generation branch, so that the trajectory-construction stage can be applied to both Dream and LLaDA-style models within a unified pipeline.

\subsubsection{D-LLM generation settings}

For each question, we generate one base answer and record the associated denoising trajectory. Dream-7B-Instruct uses 64 denoising steps, while LLaDA-8B-Instruct uses 32 steps. Both models use a maximum answer length of 32 tokens, temperature $0.0$, and top-$p$ $0.95$. The resulting runs expose 64 forward calls for Dream and 32 forward calls for LLaDA. We retain the final eight forward calls for hidden-evidence construction.

\subsubsection{Retained trajectory states and candidate positions}

Rather than storing the full hidden trajectory, we retain only a fixed number of recent denoising forwards and a sparse subset of token positions within each retained forward. In the current implementation, the reported configuration keeps the last $N=8$ forwards and allows at most 16 changed-token positions per retained forward, so that the maximum number of captured positions is
\[
\texttt{max\_M} = 1 + \texttt{max\_changed\_positions} = 17,
\]
where the extra slot corresponds to the last-token position. At each forward call, the builder compares the current input sequence with the previous one, identifies positions whose token IDs have changed, removes the last-token position from that changed set if necessary, caps the remaining changed positions, and then forms the final capture set as the union of the capped changed positions and the current last-token position. This yields a sparse but trajectory-aware set of candidate locations that emphasizes actively revised positions while always preserving a compact final-state signal.

For every retained step-layer pair, the script collects the hidden vectors at the captured positions from all decoder layers. Internally, the hook collector stores these per-layer vectors in the form $[M,H]$, where $M$ is the number of captured positions in that forward and $H$ is the original hidden dimension. The collected states are then stacked into a tensor of shape $[L,M,H]$, where $L$ is the number of retained layers. If a particular layer is missing a captured tensor for a forward, the builder pads that slot with zeros so that the packed trajectory representation remains rectangular and can be written efficiently to disk. This defensive missing-layer fallback was never triggered in the reported experiments, corresponding to a 0\% activation rate.

In addition to the trajectory features themselves, the builder writes a metadata record for each processed example to \texttt{meta.jsonl}. Each record stores the original dataset id, question text, gold answers, generated base answer, prompt and sequence lengths, generation settings, the observed number of forward calls, the retained forward IDs, the kept changed positions, the kept capture positions, the OPORP configuration, and the corresponding \texttt{feature\_row} index. The same stage stores both a hard string-match label and an LLM-judge label, together with auxiliary judge fields such as hallucination type, confidence, suspect span, and parsed output. All reported HIVE experiments use the post-processed judge correctness target for both the selector and verifier. The separately stored hard substring-match indicator is retained only as auxiliary metadata and is not used as an independent training target.

\subsubsection{OPORP compression and mmap storage format}

To keep trajectory storage tractable, we compress the collected hidden vectors using OPORP before writing them to disk. For each model layer, we instantiate a fixed OPORP mapping from the original hidden dimension $H$ to the compressed evidence dimension $r$. The mapping is determined by a layer-specific randomized permutation and projection configuration and is reused across all examples, retained denoising steps, and token positions for that layer. Projection configurations are cached and reloaded to ensure deterministic feature construction under the same $(L,H,r,\text{seed})$ setting. In the reported experiments, we use $r=64$ and store the projected hidden states in float16 format, substantially reducing storage while preserving the compressed representations used by selector training and evidence export.

The packed trajectory features are written as memory-mapped arrays under the run directory. The main tensor \texttt{features\_act.f16.mmap} has shape $[S,N,L,\texttt{max\_M},r]$, where $S$ is the number of processed examples in the current run window. The companion file \texttt{features\_meta.json} records the tensor layout, file names, dtypes, and symbolic dimensions $(S,N,L,\texttt{max\_M},r)$, allowing downstream scripts to read the stored features without hard-coding shapes. This mmap-based design is used consistently by both selector training and evidence export, and makes it possible to resume interrupted preprocessing runs while preserving row-level alignment between feature tensors and metadata.

\subsection{Step-Layer Selector Training}

After constructing compressed trajectory features, we train a step-layer-aware selector to identify the most informative hidden evidence units for downstream verification. The selector operates on the memory-mapped trajectory tensors produced in the preprocessing stage and learns, for each example, which retained step-layer pairs should be preserved under a fixed evidence budget. In our implementation, selector training is performed directly on the compressed evidence representation rather than on the original hidden states, which keeps training efficient while preserving the trajectory structure required for evidence selection.

\subsubsection{Two-stream feature construction for selector training}

The selector reads the memory-mapped evidence tensor \texttt{features\_act.f16.mmap} of shape $[S,N,L,\texttt{max\_M},r]$ together with the corresponding capture lengths \texttt{features\_cap\_len.i16.mmap}. Here, $S$ is the number of samples, $N$ is the number of retained denoising forwards, $L$ is the number of retained layers, and $r$ is the compressed evidence dimension. For each retained step, the training dataset reconstructs a two-stream feature at every layer by taking (i) the compressed hidden vector at the last captured position and (ii) the mean-pooled compressed hidden vector over the changed-token positions. Concretely, if a retained forward contains $m$ captured positions, the final slot is interpreted as the last-token feature and the first $m-1$ slots are averaged to produce the changed-token feature. The resulting per-pair representation is therefore
\[
u_{t,\ell} = [u^{\mathrm{last}}_{t,\ell} ; u^{\mathrm{chg}}_{t,\ell}] \in \mathbb{R}^{2r},
\]
and the full example is reshaped into a grid of size $N \times L$ or, equivalently, a flattened set of $T = N L$ candidate evidence units. In the current configuration, the default feature dimension is $2r = 128$ when $r=64$.

Labels for selector training are obtained from the metadata file \texttt{meta.jsonl}. In all reported experiments, the training script maps each original example ID to the post-processed judge correctness target and aligns these targets with the corresponding feature rows using \texttt{features\_row\_ids.i32.mmap}. The predefined train, validation, and test example IDs are then preserved throughout selector training and evidence export.

\subsubsection{Step-Layer-Aware Joint Gating and Top-\texorpdfstring{$K$}{K} Selection}

The selector itself is implemented as a joint gating model over all retained step-layer pairs. Given the two-stream feature tensor $X \in \mathbb{R}^{N \times L \times 2r}$ for one example, the model augments each pair feature with a learnable step embedding and a learnable layer embedding. More precisely, for each pair $(t,\ell)$, the gate input is formed as
\[
\tilde{u}_{t,\ell} = [u_{t,\ell}; a_t; b_\ell],
\]
where $a_t$ is the step embedding and $b_\ell$ is the layer embedding. A shared multilayer perceptron $f_{\mathrm{gate}}$ then produces a scalar relevance score $s_{t,\ell}=f_{\mathrm{gate}}(\tilde{u}_{t,\ell})$ for each step-layer pair, yielding a score tensor of shape $[N,L]$, or $[T]$ after flattening. This architecture makes the selector explicitly step-layer-aware while still sharing the same scoring function across all retained trajectory locations.

To obtain a fixed evidence budget, the model applies a softmax over all $T$ pair scores and then uses a straight-through top-$K$ mechanism. During training, the top-$K$ indices are selected from the flattened score vector, while the hard selection mask is combined with the soft pair probabilities so that gradients continue to propagate through the score distribution. Let $\tilde{w}_{t,\ell}$ denote the resulting straight-through selection weight for pair $(t,\ell)$. The aggregated evidence representation is
\[
m
=
\sum_{(t,\ell)}
\tilde{w}_{t,\ell}\tilde{u}_{t,\ell},
\]
where $\tilde{u}_{t,\ell}$ is the concatenated two-stream, step-embedding, and layer-embedding representation. This pooled representation is passed to a lightweight probe network that predicts the binary hallucination label for the example. Selector training is therefore driven by example-level hallucination supervision rather than a direct pair-ranking loss.

\subsubsection{Training objective, regularization, and split protocol}

The selector is trained with binary supervision using a sigmoid-based binary cross-entropy loss on the probe logits. The reported configuration additionally applies binary-label reweighting computed from the class counts in the current run directory to mitigate label imbalance. In addition to the classification loss, the implementation includes an entropy regularizer over the selected top-$K$ weights. Let $\bar{w}_1,\ldots,\bar{w}_K$ denote these weights after renormalization over the retained pairs:
\[
\mathcal{L}_{\mathrm{ent}}
=
-
\sum_{j=1}^{K}
\bar{w}_j\log\bar{w}_j.
\]
The combined training objective is
\[
\mathcal{L}
=
\mathcal{L}_{\mathrm{cls}}
+
\lambda_{\mathrm{ent}}\mathcal{L}_{\mathrm{ent}}.
\]
The entropy term penalizes diffuse top-$K$ weights and encourages a more concentrated selection distribution.

For each setting, the selector uses the predefined train, validation, and test indices described in Appendix~\ref{app:split_protocol}. The model is optimized with AdamW, and the training loop tracks validation metrics such as accuracy, F1, and AUROC. Early stopping is applied according to a configurable validation metric, and the best checkpoint is stored together with the feature shapes, split indices, and training history. In addition to the checkpoint itself, the training script writes a result JSON containing dataset statistics, hyperparameters and train/validation/test metrics.

\subsection{Evidence Export and Verifier Dataset Construction}

After training the step-layer selector, we export its selected evidence for downstream verifier training and evaluation. This stage serves as the bridge between selector-side trajectory modeling and verifier-side structured prediction: the selector identifies the top-$K$ step-layer units for each example, and the exported evidence is then repackaged into a JSON-pair format that can be consumed by the verifier language model. In our pipeline, these two steps are implemented separately as evidence export and verifier dataset packing.

\subsubsection{Exported evidence shards and stored fields}

The export stage reads the trained selector checkpoint together with the memory-mapped trajectory features and applies the selector to a chosen split. For each example, it reconstructs the same two-stream representation used during selector training, computes the selector scores, and extracts the top-$K$ step-layer pairs together with their corresponding normalized selector weights. The exported evidence is written in shard-wise \texttt{.npz} files, separately for train, validation, and test when split indices are available.

Each evidence shard stores the selected pair indices in several aligned forms, including the flattened pair index, the step IDs, the layer IDs, and the associated top-$K$ selector weights. In addition, the export script records the forward id and capture-length information associated with each selected pair, so that the selected evidence remains aligned with the original trajectory representation. When requested, the exporter also saves the selected compressed activations themselves as \texttt{act\_top}, together with a derived two-stream representation for each selected evidence unit. Optional fields such as full selector scores, normalized pair probabilities, and step- or layer-aggregated scores can also be included for analysis, but they are not required for the default verifier training path.

Overall, the exported shard format keeps the verifier interface compact: instead of exposing the full trajectory tensor, it retains only the selector-chosen evidence together with the minimal alignment information needed to reconstruct its step-layer identity and feature content.

\subsubsection{Construction of verifier JSON-pair datasets}

The packing stage converts the exported evidence into structured verifier examples. It first scans the exported evidence shards for a given split, then aligns each shard row with the corresponding metadata entry through the original example id stored in \texttt{meta.jsonl}. From that metadata record, the packer resolves the question text, the base model answer, and the same post-processed judge correctness target used during selector training, which becomes the binary verifier target.

For each example, the packer constructs a chat-style verifier input consisting of a system prompt and a user prompt. The user prompt contains the question, the base answer, and the list of selected $(\text{step}, \text{layer})$ pairs. The target is stored as a single JSON object containing a binary decision, a hallucination type, an evidence field that reproduces the selected pairs, and a short rationale. In the packed verifier format, the decision semantics are fixed as
\[
\begin{aligned}
0 &= \text{wrong/hallucinated},\\
1 &= \text{correct/non-hallucinated}.
\end{aligned}
\]
This convention is used consistently during verifier training and evaluation.

The packed examples are written as \texttt{train.jsonl}, \texttt{val.jsonl}, and \texttt{test.jsonl}. Each row contains the chat messages, the target JSON string, and an \texttt{evidence} field that stores a relative pointer to the source evidence shard together with the row index inside that shard. The evidence field also keeps the selected pair list and, when enabled, lightweight top-$K$ metadata such as selector weights and capture lengths. This design allows the verifier dataset to remain compact while preserving a direct link back to the exported evidence shards for feature loading and debugging.

Taken together, the export and packing stages transform selector outputs into a reusable verifier-training interface: the selector contributes a sparse set of informative step-layer evidence units, and the packer turns those units into structured supervision examples for verification conditioned on hidden-evidence prefixes.

\subsection{Verifier Training with Hidden-Evidence Prefixes}

After packing the verifier dataset, we train a verifier language model that combines textual question-answer context with the exported hidden evidence. In our implementation, the verifier uses Qwen2.5-7B-Instruct as the backbone language model, while the selected top-$K$ evidence units are injected through a continuous hidden-evidence prefix before the textual prompt. This design allows the verifier to condition its structured judgment not only on the final base answer, but also on the hidden evidence selected from the denoising trajectory.

\subsubsection{Hidden-Evidence Prefix Encoder}

For each packed example, the verifier dataset loads the referenced evidence shard and reconstructs the top-$K$ evidence representation. When the exported shard contains \texttt{act\_top}, the loader rebuilds the same two-stream feature used by the selector, namely a concatenation of a last-token vector and a mean-pooled changed-token vector. If only a pooled evidence tensor is available, the loader falls back to a compatible two-stream representation. In either case, the final verifier-side evidence input has shape $[K, 2r]$ and is accompanied by the corresponding selector weights, step IDs, and layer IDs.

These top-$K$ evidence vectors are mapped into the verifier hidden space by the hidden-evidence prefix encoder. For each selected unit, the encoder projects the two-stream evidence vector into the verifier hidden dimension and fuses the projected representation with embeddings derived from its step identity, layer identity, and normalized selector weight. All three side signals are enabled in the reported configuration. The resulting evidence embeddings have shape $[K,d_{\mathrm{model}}]$ and form the continuous hidden-evidence prefix supplied to the verifier.

\subsubsection{Prefix-conditioned Qwen verifier}

The verifier wrapper prepends the hidden-evidence prefix embeddings to the ordinary textual token embeddings before passing the combined sequence to the base Qwen model. Concretely, the wrapper first tokenizes the verifier prompt built from the packed chat messages, then concatenates the hidden-evidence prefix embeddings with the textual token embeddings, and finally extends the attention mask so that the model attends jointly to both the evidence prefix and the text tokens. During training, the prefix positions are excluded from the language-modeling loss by setting their labels to \(-100\), so supervision is applied only to the generated structured target sequence.

This wrapper is used in both training and inference. At training time, the prefix is injected only once at the beginning of the sequence; during autoregressive decoding, subsequent generation steps reuse the cached decoder state and do not repeatedly reinsert the prefix. In this way, the verifier behaves like a standard causal language model whose initial hidden context has been augmented with the selected trajectory evidence.

\subsubsection{Training data format and supervision targets}

Each verifier example consists of a system message, a user message containing the question, base answer, and selected step-layer pairs, and a target JSON string. The supervision target includes a binary decision field, a hallucination-type field, an evidence field reproducing the provided step-layer pairs, and a short rationale. The dataset tokenizes the prompt and target jointly, masks out the prompt portion in the labels, and trains the verifier to autoregressively generate only the structured JSON output.

In the reported configuration, verifier training uses Hugging Face training utilities with LoRA adaptation and bfloat16 mixed precision. The data collator pads the text sequence and batches the evidence tensors \((x_{\mathrm{top}}, w_{\mathrm{top}}, \texttt{step\_ids}, \texttt{layer\_ids})\) in parallel, so that each training batch contains both the textual supervision signal and the aligned hidden-evidence prefix inputs. This yields a verifier that is trained directly on evidence-conditioned structured prediction rather than on scalar classification alone.

\subsection{Evaluation Protocol and Decision-Logit Scoring}

We evaluate the trained verifier from two complementary perspectives: structured-output quality and scalar hallucination detection quality. At evaluation time, the verifier reads the packed JSONL examples together with their referenced evidence shards, reconstructs the top-$K$ hidden evidence used to form the verifier prefix, and generates a structured JSON prediction under greedy decoding. The same evaluation pipeline is used across train, validation, and test splits, with the main reported results taken from held-out evaluation splits.

\subsubsection{Structured-output parsing and field-level evaluation}

For each example, the evaluation script first rebuilds the verifier prompt from the stored chat messages and reloads the corresponding top-$K$ evidence from the exported shard. The verifier then generates a structured JSON prediction conditioned on both the textual prompt and the hidden-evidence prefix.

The generated text is parsed by extracting the first valid JSON object from the decoder output. Structured-output metrics are then computed by comparing the predicted JSON fields with the target JSON stored in the packed dataset. In particular, the evaluation script tracks valid-JSON rate, decision accuracy, hallucination-type accuracy, evidence-pair reproduction accuracy, rationale-reproduction accuracy, and full-schema exact match across all structured fields. This design allows us to evaluate HIVE not only as a detector, but also as a verifier that emits schema-constrained structured outputs.

\subsubsection{Decision-emission logit extraction}

In addition to the structured JSON itself, we derive a continuous hallucination score from the verifier's decision-emission logits. During greedy decoding, the evaluation script monitors the generated output until the \texttt{"decision"} field is reached. At the step where the model emits the binary decision token, the script collects the logits associated with tokenizations corresponding to the digits \texttt{0} and \texttt{1}, aggregates them into a binary logit contrast, and converts the result into a probability \(p_1\) for the event
\[
\texttt{decision} = 1,
\]
where \(1\) denotes a correct/non-hallucinated answer in the packed verifier format. The final hallucination score is then defined as
\[
s_{\mathrm{hall}} = 1 - p_1.
\]
This score preserves finer-grained verifier confidence than the hard decision alone and serves as the scalar signal used for threshold-free hallucination detection evaluation.

\subsubsection{Threshold selection and transferred evaluation}

Given the continuous hallucination score, we compute threshold-free metrics such as AUROC and AUPRC directly from the full score distribution. For thresholded evaluation, the implementation supports several threshold-selection strategies, including best-F1 and Youden-style selection. In the main experiments, thresholds are selected on the validation split and then transferred unchanged to the test split, so that thresholded metrics reflect a realistic deployment-style setting rather than per-split re-optimization. Once a threshold is fixed, the evaluation script reports precision, recall, F1, accuracy, balanced accuracy, specificity, and the associated confusion matrix.

Taken together, this evaluation protocol yields two complementary views of verifier quality. The structured-output metrics assess whether the model generates valid
and schema-constrained diagnostic fields, while the decision-logit score provides a continuous hallucination signal for standard detection metrics and downstream thresholded classification.

\subsection{Training Hyperparameters and Runtime Configuration}

We summarize the main training and runtime settings in Tables~\ref{tab:appendix_selector_hp} and~\ref{tab:appendix_verifier_hp}. Table~\ref{tab:appendix_selector_hp} reports the default selector-training configuration used for step-layer-aware two-stream selection, while Table~\ref{tab:appendix_verifier_hp} summarizes the verifier training and decoding settings used for structured verification conditioned on hidden-evidence prefixes. Unless otherwise specified, the experiments in the paper follow these configurations, with model-specific layer counts inherited from the underlying D-LLM.

\begin{table}[t]
\centering
\footnotesize
\setlength{\tabcolsep}{5pt}
\renewcommand{\arraystretch}{1.02}
\begin{tabularx}{\columnwidth}{@{}>{\raggedright\arraybackslash}p{0.44\columnwidth}>{\raggedright\arraybackslash}X@{}}
\toprule
\textbf{Parameter} & \textbf{Value} \\
\midrule
Optimizer & AdamW \\
Batch size & 128 \\
Epochs & 30 \\
Learning rate & \(1\times10^{-3}\) \\
Weight decay & \(1\times10^{-4}\) \\
Dropout & 0.2 \\
Train / val / test split & 0.7 / 0.1 / 0.2 \\
Top-\(K\) budget \(K\) & 16 \\
Step embedding dim & 16 \\
Layer embedding dim & 16 \\
Gate input dim & 160 \\
Gate hidden dim & 128 \\
Probe hidden dim & 512 \\
Gate temperature \(\tau\) & 1.0 \\
Entropy regularization \(\lambda_{\mathrm{ent}}\) & \(1\times10^{-3}\) \\
Binary-label reweighting & Enabled \\
Early-stop patience & 10 \\
Early-stop metric & AUROC \\
Random seed & 42 \\
\bottomrule
\end{tabularx}

\caption{\textbf{Selector training hyperparameters.}
Default configuration of the step-layer-aware two-stream selector
used in the evidence-selection stage.}
\label{tab:appendix_selector_hp}
\end{table}

\begin{table}[t]
\centering
\footnotesize
\setlength{\tabcolsep}{4pt}
\renewcommand{\arraystretch}{1.02}
\begin{tabularx}{\columnwidth}{@{}>{\raggedright\arraybackslash}p{0.43\columnwidth}>{\raggedright\arraybackslash}X@{}}
\toprule
\textbf{Parameter} & \textbf{Value} \\
\midrule
Verifier backbone & Qwen2.5-7B-Instruct \\
Retained steps \(N\) & 8 \\
Compressed dimension \(r\) & 64 \\
Top-\(K\) evidence budget & 16 \\
Maximum input length & 1024 \\
Padding multiple & 8 \\
Train / val / test split & 0.7 / 0.1 / 0.2 \\
Batch size & 1 \\
Gradient accumulation & 8 \\
Epochs & 4 \\
Learning rate & \(2\times10^{-5}\) \\
Weight decay & 0.0 \\
Warmup ratio & 0.03 \\
Logging steps & 10 \\
Save steps & 500 \\
Eval steps & 500 \\
Training dtype & bfloat16 \\
LoRA & Enabled \\
LoRA rank / alpha / dropout & 8 / 16 / 0.05 \\
\midrule
Maximum decoding length & 256 \\
Inference dtype & bfloat16 \\
Threshold selection & Validation best-F1, transferred unchanged to test \\
\bottomrule
\end{tabularx}
\caption{\textbf{Verifier training and decoding hyperparameters.} Default configuration of hidden-evidence prefix verifier training and structured evaluation.}
\label{tab:appendix_verifier_hp}
\end{table}

\subsection{Baseline Implementation Details}
\label{app:baseline_impl}

\paragraph{Perplexity and LN-Entropy.}
For Dream-7B-Instruct, Perplexity is computed by teacher forcing on the generated answer tokens and exponentiating the mean answer-token negative log-likelihood. For LLaDA-8B-Instruct, which does not expose a standard autoregressive likelihood, we compute pseudo-perplexity through single-token masked reconstruction over the answer tokens and exponentiate the resulting mean negative log-likelihood. LN-Entropy is computed as the mean predictive entropy over answer-token positions. Both methods use a maximum input length of 1024 tokens.

\paragraph{Sampling-based uncertainty baselines.}
Semantic Entropy, Lexical Similarity, and EigenScore use 10 independently sampled answers per question with temperature $0.8$, top-$p$ $0.95$, and a maximum generation length of 32 tokens. For LLaDA sampling, we use low-confidence remasking. Semantic Entropy computes entropy over semantic clusters formed from the sampled answers. Lexical Similarity is implemented as one minus the mean pairwise token-level Jaccard similarity across the sampled answers. EigenScore mean-pools the answer representations, applies $\ell_2$ normalization, constructs a Gram-matrix covariance surrogate, and computes the sum of the log eigenvalues with numerical constant $\epsilon=10^{-6}$.

\paragraph{Latent and trajectory baselines.}
CCS and TSV operate on mean-pooled, $\ell_2$-normalized answer representations from the target D-LLM. The CCS probe uses a hidden dimension of 512 and dropout 0.1 and is trained for 10 epochs using AdamW with batch size 256, learning rate $10^{-3}$, and weight decay $10^{-4}$. TraceDet is implemented over the same eight retained trajectory steps used by HIVE. At each step, it concatenates the layer-averaged last-token feature and mean changed-token feature, yielding a 128-dimensional step representation. The resulting trajectory is step-normalized and dimension-wise standardized before being processed by a one-layer trajectory encoder. TraceDet is trained for 20 epochs with batch size 256, learning rate $10^{-3}$, weight decay $10^{-4}$, dropout 0.1, and random seed 42.

\paragraph{Text-only verifier control.}
For each model--benchmark setting, we separately train a verifier-backbone-matched text-only control using Qwen2.5-7B-Instruct. The control uses the same primary correctness labels, question-level train/validation/test assignments, LoRA rank, alpha, and dropout, optimizer configuration, learning-rate schedule, checkpoint-selection procedure, maximum input and decoding lengths, and decision-logit scoring protocol as the HIVE verifier. Its input contains only the question and the target D-LLM's generated answer. The hidden-evidence encoder, continuous evidence prefix, selector weights, evidence tensors, and step-layer identifiers are omitted. Threshold-free metrics are computed from the decision-emission logits, and thresholded predictions use a validation-selected best-F1 threshold transferred unchanged to the test split.

\paragraph{Threshold selection.}
For every baseline producing a continuous hallucination score, higher values indicate greater hallucination likelihood. Threshold-free metrics are computed directly from these scores. For thresholded metrics, the decision threshold is selected on the validation split by maximizing F1 and is then applied unchanged to the corresponding test split.

\section{Automatic Label Construction}
\label{app:judge_supervision}

\subsection{Judge Model and Decoding}

We use Qwen2.5-7B-Instruct as the automatic judge that constructs the supervision labels used by both the step-layer selector and the evidence-conditioned verifier. For each example, the judge receives the original question, all available gold-answer variants, and the answer generated by the target D-LLM. Decoding is deterministic, using greedy generation with sampling disabled and a maximum output length of 160 tokens. The judge assigns label 1 only when the generated answer is semantically equivalent to at least one valid gold answer. Answers that are incorrect or whose correctness cannot be established receive label 0.

\subsection{Complete Judge Prompt}

The following reproduces the exact system and user message content used by the label-construction scripts for both Dream and LLaDA. Tokenizer-specific chat-template control tokens are omitted. The bracketed placeholders are replaced by the question, the JSON-serialized gold-answer list, and the target D-LLM answer for each example.

\begin{Verbatim}[breaklines=true,breakanywhere=true,fontsize=\small]
[SYSTEM]
You are a strict grader for question answering.
Task: Decide whether the MODEL_ANSWER is correct given the QUESTION and GOLD_ANSWERS.
Output ONLY a JSON object with keys: label, confidence, hallucination_type, suspect_span, rationale.

Label rule:
- label=1 if MODEL_ANSWER is semantically equivalent to ANY gold answer (treat gold list as acceptable answers/aliases).
- Otherwise label=0.
- If unsure, label=0.

Hallucination type taxonomy (choose exactly one):
- none: answer is correct.
- incomplete: missing key entity/number/date required by the question.
- irrelevant: answer is about a different topic.
- fabrication: invents non-existent entities/papers/laws/quotes.
- reasoning_error: wrong conclusion due to invalid reasoning.
- factual_error: clear contradiction with GOLD_ANSWERS.
- unsupported: ONLY if GOLD_ANSWERS are empty/insufficient to verify.

Priority rules:
1) If label=1 -> hallucination_type MUST be 'none'.
2) Else if contradiction -> 'factual_error'.
3) Else if missing key -> 'incomplete'.
4) Use 'unsupported' only when genuinely cannot verify.

suspect_span rules:
- If label=1: suspect_span MUST be "".
- If label=0: suspect_span MUST be a VERBATIM substring from MODEL_ANSWER, minimal, <= 80 chars.
- Do NOT paraphrase.

[USER]
QUESTION: <QUESTION>
GOLD_ANSWERS: <GOLD_ANSWERS>
MODEL_ANSWER: <MODEL_ANSWER>
Return JSON now.
\end{Verbatim}

\subsection{Parsed Fields and Label Semantics}

The judge is required to return a single JSON object containing five fields: \texttt{label}, \texttt{confidence}, \texttt{hallucination\_type}, \texttt{suspect\_span}, and \texttt{rationale}. The binary \texttt{label} field provides the judge's initial correctness decision, with 1 denoting a correct answer and 0 denoting an incorrect or unverified answer. This initial decision is subsequently normalized and may be revised by the deterministic consistency checks described below. The resulting post-processed correctness label, rather than the raw parsed field alone, is used as the primary binary supervision target for both stages of HIVE. The \texttt{confidence} field records the judge's confidence in its initial binary decision. The \texttt{hallucination\_type} field provides an auxiliary error category, the \texttt{suspect\_span} field identifies a minimal verbatim substring of an incorrect generated answer, and the \texttt{rationale} field contains a short textual justification. These auxiliary fields are retained as metadata and are used, where applicable, to construct structured verifier targets and support diagnostic analysis.

\subsection{Deterministic Post-checks and Failure Handling}

The parsed judge output is normalized and subjected to deterministic consistency checks before being stored. If the final label is 1, the hallucination type is forced to \texttt{none} and the suspect span is cleared. If the final label is 0, the suspect span must be a verbatim substring of the generated answer; invalid or non-verbatim spans are sanitized. Hallucination types that are missing, inconsistent with the binary label, or outside the allowed taxonomy are normalized to a valid category. Questions requiring multiple answer elements are checked for sufficient gold-answer coverage, and strict or soft gold-answer evidence may revise a judge decision or append a post-check explanation when the generated fields are inconsistent with deterministic matching evidence.

If no valid JSON object can be parsed, the example is conservatively assigned label 0. The fallback hallucination type is \texttt{incomplete} when gold answers are available and \texttt{unsupported} otherwise, and the suspect span is left empty. This fallback prevents malformed judge generations from being treated as correct supervision.

\subsection{Raw Label-Construction Example}

The following example shows the example-specific content supplied to the automatic judge, its raw generated JSON, and the final metadata fields after deterministic post-processing.

\begin{Verbatim}[breaklines=true,breakanywhere=true,fontsize=\small]
QUESTION:
Which magazine was started first Arthur's Magazine or First for Women?

GOLD ANSWERS:
["[Arthur's Magazine]"]

MODEL ANSWER:
Arthur's Magazine

HARD STRING-MATCH LABEL:
1

RAW JUDGE OUTPUT:
{
  "label": 1,
  "confidence": 1,
  "hallucination_type": "none",
  "suspect_span": "",
  "rationale": "The model answer matches one of the gold answers."
}

FINAL STORED FIELDS:
{
  "label_llm_judge": 1,
  "judge_hallucination_type": "none",
  "judge_confidence": 1.0,
  "judge_suspect_span": "",
  "judge_suspect_valid": true,
  "judge_parsed": {
    "label": 1,
    "confidence": 1.0,
    "hallucination_type": "none",
    "suspect_span": "",
    "rationale": "The model answer matches one of the gold answers."
  }
}
\end{Verbatim}

This example shows the raw judge generation and the corresponding normalized metadata stored by the preprocessing pipeline. The post-processed judge label, rather than the separately stored hard substring-match indicator, is used as the binary target for both selector and verifier training. For correct examples, post-processing enforces the consistency constraints that the hallucination type is \texttt{none} and the suspect span is empty.

\subsection{Relationship to Hard String Matching}

The preprocessing stage separately stores a hard substring-match indicator as auxiliary metadata. This indicator is not used as an independent vote, an additional training target, or a substitute for semantic judging. All reported experiments use the post-processed judge correctness target for selector and verifier training. That target is initialized from the parsed LLM-judge output and may be revised by deterministic gold-answer consistency checks described above. These checks operate directly on the question, gold answers, and generated response and should not be confused with combining the separately stored hard-match indicator with the judge label. The auxiliary hard-match indicator may disagree with the final target because the presence of a gold-answer substring does not guarantee that the complete generated statement is semantically correct.

\section{Human Annotation Audit Details}
\label{app:human_audit}

\subsection{Sampling and Annotation Protocol}

We audit 120 examples sampled from all six model--dataset settings. The audit set was fixed before annotation and preserved with the original example identifier, target D-LLM, benchmark, question, gold-answer variants, generated answer, post-processed automatic binary label, and post-processed automatic hallucination type. We sampled 20 examples from each of the six model--dataset settings, yielding 120 examples in total. Within each setting, sampling was stratified by the post-processed automatic binary label, with 10 judge-labeled correct examples and 10 judge-labeled incorrect examples drawn from the held-out test split. The automatic labels were used only for sampling stratification and were hidden from the annotators during annotation.

Two annotators independently labeled every example while blinded to the automatic judge label, hallucination type, confidence, suspect span, and rationale. For binary correctness, annotators assigned 1 only when the generated answer was semantically correct with respect to the question and at least one acceptable gold answer; otherwise they assigned 0. For hallucination type, annotators used the same seven-category post-processed taxonomy as the automatic label-construction pipeline, assigning \texttt{none} to correct answers and one of \texttt{factual\_error}, \texttt{fabrication}, \texttt{unsupported}, \texttt{incomplete}, \texttt{irrelevant}, or \texttt{reasoning\_error} to incorrect answers.

Inter-annotator agreement and Cohen's $\kappa$ were computed from the two independent annotation files before disagreement resolution. Disagreements were resolved through discussion between the two annotators until consensus, yielding the final human audit labels used for the auto--human accuracy and macro-F1 calculations. The automatic side of this comparison uses the post-processed judge label and type rather than the raw unnormalized judge generation. Because the audit set is balanced by the automatic binary label within each setting, the reported auto--human accuracy and macro-F1 characterize this controlled audit sample rather than the natural label prevalence of the full datasets.

\subsection{Aggregate Results}

\begin{table}[t]
\centering
\footnotesize
\setlength{\tabcolsep}{2.5pt}
\renewcommand{\arraystretch}{1.05}
\begin{tabularx}{\columnwidth}{@{}>{\raggedright\arraybackslash}Xrrrr@{}}
\toprule
\textbf{Label task} & \shortstack{\textbf{Annotator}\\\textbf{agreement}} & \textbf{$\kappa$} & \shortstack{\textbf{Auto--human}\\\textbf{Acc.}} & \textbf{Macro-F1} \\
\midrule
Binary correctness & 94.17 & 0.883 & 92.50 & 92.46 \\
Hallucination type & 83.33 & 0.799 & 78.33 & 80.82 \\
\bottomrule
\end{tabularx}
\caption{\textbf{Human audit of automatic supervision on a balanced 120-example set.} Agreement and auto--human metrics are reported in percent; Cohen's $\kappa$ is reported on its original scale.}
\label{tab:human_label_audit}
\end{table}

Across the 120 audited examples, binary correctness reaches 94.17\% inter-annotator agreement with Cohen's $\kappa=0.883$. Relative to the final human audit labels, the automatic binary labels achieve 92.50\% accuracy and 92.46 macro-F1. Hallucination-type annotation is less reliable, with 83.33\% inter-annotator agreement and Cohen's $\kappa=0.799$; the automatic type labels achieve 78.33\% accuracy and 80.82 macro-F1. These results support the reliability of the primary binary supervision while showing that the fine-grained type labels contain appreciably more noise.

\subsection{Scope of the Audit}

The audit evaluates the quality of the automatically constructed binary and type labels. It does not validate the selected step-layer units as human-readable evidence, and it does not establish that the verifier-generated rationale is faithful to the hidden trajectory. Accordingly, binary correctness is treated as the primary task, while hallucination type and rationale are retained as auxiliary diagnostic outputs.

\section{Prompt Templates and Output Schema}
\label{app:prompt_schema}

\subsection{Text-Only Control Prompt}

The verifier-backbone-matched text-only control receives only the original question and the target D-LLM's generated answer. Its prompt omits the hidden-evidence section, selected step-layer pair list, evidence-shard references, selector weights, and all continuous prefix inputs. The control uses the same decision semantics and decision-logit scoring procedure as HIVE, thereby isolating the contribution of trajectory-derived evidence from ordinary textual verification.

\subsection{Verifier Prompt Format}

HIVE uses a chat-style verifier prompt that combines ordinary textual context with an explicit schema instruction. Each verifier example is organized as a two-message conversation consisting of a system message and a user message. The system message defines the model as a verifier, explains that it should judge whether the base answer is wrong or hallucinated, and states that hidden evidence has already been injected through hidden-evidence prefix embeddings. It also requires the model to output a single JSON object and forbids any extra natural-language text outside the schema.

The user message contains four main parts. First, it states the task, namely to judge whether the base answer is hallucinated with respect to the question. Second, it provides the original question. Third, it includes the base model answer to be verified. Fourth, it presents the selected hidden evidence in symbolic form as a list of top-$K$ \((\text{step}, \text{layer})\) pairs. In other words, although the verifier does not directly read the full denoising trajectory in text form, the prompt still exposes the identities of the selected evidence units that correspond to the injected hidden-evidence prefixes.

The prompt additionally includes an explicit output-schema specification. This schema tells the verifier that it must produce a JSON object containing a binary decision field, a hallucination-type field, an evidence field that reproduces the provided step-layer pairs, and a short rationale. In the packed verifier format, the decision semantics are fixed as
\[
\begin{aligned}
0 &= \text{wrong/hallucinated},\\
1 &= \text{correct/non-hallucinated}.
\end{aligned}
\]
This convention is stated directly in the prompt so that the verifier is trained and evaluated under a consistent decision interface.

Overall, the verifier prompt is designed to do two things simultaneously: it provides the usual textual verification context through the question and base answer, and it anchors the hidden-evidence interface by exposing the selected step-layer pairs and the required structured output format. This makes the verifier behave as a schema-constrained evidence-conditioned judge rather than as a free-form text generator.

\subsection{Output JSON Schema}

The verifier is trained to generate a single JSON object with a fixed schema. This schema is explicitly specified in the verifier prompt and is also used as the supervision target during training. In our implementation, each target JSON contains four top-level components: a binary decision field, a hallucination-type field, an evidence field, and a short rationale. The evidence field itself contains a single subfield, \texttt{pairs}, which reproduces the provided top-$K$ \((\text{step}, \text{layer})\) indices exactly.

The schema can be summarized as follows:
\begin{Verbatim}[breaklines=true,breakanywhere=true,fontsize=\small]
{
  "decision": 0 or 1,
  "hallucination_type": one of
    ["none", "factual_error", "fabrication", "unsupported",
    "incomplete", "irrelevant", "reasoning_error"],
  "evidence": {
    "pairs": the provided (step, layer) pairs
  },
  "rationale": a single short sentence
}
\end{Verbatim}

Among these fields, \texttt{decision} is the primary binary verification output. In the packed verifier format, the decision semantics are fixed as
\[
\begin{aligned}
0 &= \text{wrong/hallucinated},\\
1 &= \text{correct/non-hallucinated}.
\end{aligned}
\]
The \texttt{hallucination\_type} field refines this binary judgment into a more specific error category. The \texttt{evidence.pairs} field reports the selected top-$K$ step-layer pairs into the output, and \texttt{rationale} provides a short natural-language justification.

This schema is used consistently throughout the pipeline. During dataset packing, the target JSON is serialized into the \texttt{target} field of each verifier example. During training, the verifier learns to autoregressively generate this structured object. During evaluation, the same schema is used for field-level comparison, including decision accuracy, hallucination-type accuracy, evidence-pair reproduction accuracy, rationale-reproduction accuracy, and full-schema exact match.

Overall, the output schema serves two purposes at once: it provides a standard binary hallucination decision for detection, and it exposes a richer verification interface that makes HIVE's predictions more structured and easier to inspect than a single scalar score alone.

\subsection{Hallucination Type Taxonomy}

In addition to the binary decision field, HIVE predicts a hallucination-type label that refines the verifier output into a more specific error category. In the current schema, the allowed type set is
\{\texttt{none}, \texttt{factual\_error}, \texttt{fabrication}, \texttt{unsupported}, \texttt{incomplete}, \texttt{irrelevant}, \texttt{reasoning\_error}\}.
These type labels are specified directly in the verifier prompt and are also used as the supervision targets during training.

The label \texttt{none} is reserved for non-hallucinated answers, namely cases in which the verifier judges the base answer to be correct. All remaining labels correspond to different kinds of hallucinated or otherwise unreliable answers. Among them, \texttt{factual\_error} refers to answers that contain a concrete incorrect fact, entity, number, or relation; \texttt{fabrication} refers to answers that invent unsupported content more substantially; \texttt{unsupported} is reserved for cases in which the available gold answers are empty or insufficient to verify the generated answer; \texttt{incomplete} refers to answers that are partially correct but omit an important required element; \texttt{irrelevant} refers to answers that do not address the question; and \texttt{reasoning\_error} refers to answers whose final conclusion is wrong because of an incorrect reasoning process.

This taxonomy provides a lightweight structured decomposition of verifier outputs rather than an exhaustive or human-gold theory of hallucination. The supervision targets are derived from the automatic judge, and the human audit in Appendix~\ref{app:human_audit} shows that type labels have lower human agreement and lower auto--human alignment than binary correctness labels. We therefore report hallucination-type accuracy separately as an auxiliary diagnostic metric and do not use it as evidence of faithful interpretability.

In the packed verifier format, the hallucination type must remain consistent with the binary decision. In particular, when \texttt{decision} \(=1\), the corresponding type should be \texttt{none}; otherwise the type should belong to one of the hallucination categories listed above. This consistency is enforced during dataset construction and checked again during structured-output evaluation.

\subsection{Verifier Dataset Example}

Each verifier example is stored as one JSONL record containing the chat-style prompt, the structured target JSON, and the evidence pointer needed to reload the exported top-$K$ evidence shard. For readability, we show below a simplified example that keeps only the fields most relevant to verifier supervision and omits auxiliary bookkeeping fields unrelated to the prompt or target format.

\begin{Verbatim}[breaklines=true,breakanywhere=true,fontsize=\small]
{
  "messages": [
    {
      "role": "system",
      "content": "You are a verifier model. Given a question and a base
      model answer, decide whether the answer is wrong/hallucinated.
      ... You MUST output a single JSON object ..."
    },
    {
      "role": "user",
      "content": "[Task]
      Judge whether the Base Answer is 
      wrong/hallucinated with respect to the Question.

      [Question]
      Who was the beautiful sister of the 
      twins Castor and Pollux and mother to Hermione?

      [Base Answer]
      The beautiful sister of the twins Castor 
      and Pollux, and mother to Hermione, 
      was Leda (or Leta) in Greek mythology.

      [Evidence]
      You received K=16 ACT token-groups 
      as hidden evidence. The (step, layer) 
      pairs of these ACT token-groups are:
      [[2, 27], [3, 27], ..., [0, 24]]

      [Output JSON Schema]
      {
        "decision": 0 or 1,
        "hallucination_type": one of ["none","factual_error","fabrication",
        "unsupported", "incomplete","irrelevant",
        "reasoning_error"],
        "evidence": {
          "pairs": the provided (step, layer) pairs
        },
        "rationale": a single short sentence
      }
      Return only the JSON."}],
  "target": {
    "decision": 0,
    "hallucination_type": "factual_error",
    "evidence": {
      "pairs": [[2, 27], [3, 27], ..., [0, 24]]
    },
    "rationale": "Based on the provided hidden 
    evidence, the base answer is likely 
    hallucinated; the suspected erroneous 
    span is: 'Leda (or Leta)'."
  }
}
\end{Verbatim}

This example illustrates the basic verifier supervision format. The textual prompt provides the question, the base answer, and the symbolic identities of the selected evidence units, while the target specifies the structured diagnostic output that the model must generate. In the packed dataset, each record also includes an \texttt{evidence} field containing a relative pointer to the source evidence shard and the row index within that shard, so that the corresponding hidden-evidence prefix features can be loaded during training and evaluation.

Two aspects of this format are particularly important. First, the target explicitly reproduces the selected step-layer pairs in the \texttt{evidence.pairs} field, which ties the generated verification output back to the hidden-evidence interface. Second, the rationale remains short and schema-constrained rather than open-ended, so the verifier is trained to emit compact verification summaries instead of unrestricted explanations. This keeps the structured output compact and schema-constrained and makes field-level evaluation straightforward.

\subsection{Structured Output Parsing and Evaluation Rules}

At inference time, the verifier generates a text sequence that is expected to contain exactly one structured JSON object. The evaluation pipeline first decodes the generated tokens into text and then extracts the first valid JSON object from the output. If no valid JSON object can be recovered, the prediction is treated as invalid for structured evaluation. This gives rise to the valid-JSON metric reported in the main paper, which measures whether the verifier successfully follows the required output format.

Once a valid JSON object is obtained, the prediction is compared against the target JSON stored in the packed dataset. The structured evaluation focuses on four fields: \texttt{decision}, \texttt{hallucination\_type}, \texttt{evidence.pairs}, and \texttt{rationale}. The \texttt{decision} field is evaluated by exact agreement with the target binary label, the \texttt{hallucination\_type} field is evaluated by exact agreement with the target type label, and the \texttt{rationale} field is evaluated by exact string match. For the \texttt{evidence.pairs} field, the prediction must reproduce the provided top-$K$ \((\text{step}, \text{layer})\) pairs exactly and in order. This metric evaluates schema following and pair reproduction rather than the semantic quality or causal importance of the selected hidden units. 

Based on these field-level comparisons, we report valid-JSON rate, decision accuracy, hallucination-type accuracy, and full-schema exact match in the structured-results table. Evidence-pair and rationale reproduction are retained as diagnostic field-level checks. The full-schema exact-match metric is the strictest: it requires all evaluated structured fields to match the target simultaneously. Rationale reproduction measures agreement with the automatically
constructed target text and should not be interpreted as an evaluation of explanation faithfulness. This makes it possible to distinguish between partial success, such as producing a correct decision but an incorrect rationale, and fully correct structured verification output.

An important aspect of this evaluation protocol is that the structured prediction is assessed independently of the continuous hallucination score used for AUROC and AUPRC. In other words, structured-output evaluation assesses whether the verifier produces a valid and schema-compliant diagnostic JSON prediction, while scalar detection evaluation measures how well the model ranks hallucinated and non-hallucinated answers. These two views are complementary, and together they capture both the detection quality and the verification quality of HIVE.

\section{Additional Data and Component Results}
\label{app:additional_results}

\subsection{Dataset Statistics}

\begin{table*}[t]
\centering
\small
\setlength{\tabcolsep}{5pt}
\begin{tabular}{llrrrr}
\toprule
\textbf{Model} & \textbf{Dataset} & \textbf{Train} & \textbf{Val} & \textbf{Test} & \textbf{Hallucination Rate (\%)} \\
\midrule
Dream-7B-Instruct & HotpotQA & 62,275 & 8,897 & 17,794 & 77.51 \\
Dream-7B-Instruct & NQOpenLike & 6,130 & 876 & 1,752 & 82.58 \\
Dream-7B-Instruct & TriviaQA & 48,902 & 6,987 & 13,973 & 61.40 \\
LLaDA-8B-Instruct & HotpotQA & 62,275 & 8,897 & 17,794 & 75.43 \\
LLaDA-8B-Instruct & NQOpenLike & 6,130 & 876 & 1,752 & 82.84 \\
LLaDA-8B-Instruct & TriviaQA & 48,902 & 6,987 & 13,973 & 64.26 \\
\bottomrule
\end{tabular}
\caption{\textbf{Dataset statistics for the six independently constructed settings.} Hallucination rate denotes the percentage of examples assigned an incorrect label by the post-processed judge target.}
\label{tab:data_statistics}
\end{table*}

\subsection{Matched Text-Only Verifier Comparison}

\begin{table*}[t]
\centering
\small
\setlength{\tabcolsep}{7pt}
\renewcommand{\arraystretch}{1.05}
\begin{tabular}{llccc}
\toprule
\textbf{D-LLM} & \textbf{Benchmark}
& \textbf{Text-only verifier}
& \textbf{Full HIVE}
& \textbf{Gain} \\
\midrule
Dream-7B-Instruct & HotpotQA
& 89.42 / 96.27 & \textbf{91.76 / 97.37}
& \textbf{+2.34 / +1.10} \\

Dream-7B-Instruct & NQOpenLike
& 83.61 / 95.39 & \textbf{85.74 / 96.60}
& \textbf{+2.13 / +1.21} \\

Dream-7B-Instruct & TriviaQA
& 88.54 / 91.75 & \textbf{92.36 / 95.37}
& \textbf{+3.82 / +3.62} \\
\midrule
LLaDA-8B-Instruct & HotpotQA
& 89.21 / 94.73 & \textbf{90.94 / 96.48}
& \textbf{+1.73 / +1.75} \\

LLaDA-8B-Instruct & NQOpenLike
& 81.11 / 94.05 & \textbf{85.71 / 96.95}
& \textbf{+4.60 / +2.90} \\

LLaDA-8B-Instruct & TriviaQA
& 86.40 / 90.90 & \textbf{90.66 / 93.98}
& \textbf{+4.26 / +3.08} \\
\midrule
\multicolumn{2}{l}{\textbf{Average gain}}
& -- & -- & \textbf{+3.15 / +2.28} \\
\bottomrule
\end{tabular}
\caption{\textbf{Verifier-backbone-matched comparison between text-only verification and HIVE.}
Each cell reports AUROC / AUPRC in percent.
The two methods use the same verifier backbone, primary correctness supervision,
question-level splits, verifier-side LoRA and optimization configuration,
checkpoint-selection procedure, and decision-logit scoring protocol.
The text-only verifier omits all hidden-evidence and step-layer inputs.
The gain column reports Full HIVE minus the text-only verifier, and the final row reports the unweighted mean gain across the six settings.}
\label{tab:text_only_vs_hive}
\end{table*}

\subsection{Detailed Input-Source Decomposition}

\begin{table}[t]
\centering
\footnotesize
\setlength{\tabcolsep}{2pt}
\renewcommand{\arraystretch}{1.05}
\begin{tabular}{@{}llrrr@{}}
\toprule
\textbf{Method} & \textbf{Input} & \textbf{AUROC} & \textbf{AUPRC} & \textbf{F1} \\
\midrule
Zero-shot Qwen & Text & 76.39 & 84.11 & 73.76 \\
Selector probe & Hidden & 84.77 & 89.84 & 80.37 \\
Text-only verifier & Text & 88.54 & 91.75 & 84.25 \\
\textbf{Full HIVE} & Text + hidden & \textbf{92.36} & \textbf{95.37} & \textbf{87.76} \\
\bottomrule
\end{tabular}
\caption{\textbf{Detailed input-source decomposition on Dream-7B-Instruct + TriviaQA.}
Zero-shot Qwen and the text-only verifier use textual input, the selector probe uses hidden trajectory features only, and Full HIVE combines textual and selected hidden evidence. Results are percentages.}
\label{tab:appendix_component_comparison}
\end{table}

\section{Detailed Ablation Analysis}
\label{app:detailed_ablations}

\subsection{Hidden-Evidence Prefix Conditioning}

We compare Full HIVE against a plain verifier retrained without evidence prefixes, the prefix-trained verifier evaluated without a prefix, and a shuffled-prefix variant that preserves prefix size while breaking evidence alignment. AUROC decreases from 92.36 to 88.54, 86.64, and 88.57, respectively, showing that the gain cannot be attributed only to the verifier backbone or to adding extra prefix tokens.

\subsection{Learned Step-Layer Evidence Selection}

We compare the learned selector with final-step-only selection, uniform allocation over steps, and random $K$-pair selection under the same evidence budget. AUROC drops from 92.36 to 89.17, 88.34, and 85.85, respectively. Together with the aggregate selection pattern in main-text Figure~\ref{fig:agg_step_layer_pattern}, these results show that useful evidence is distributed across the trajectory but is not uniformly informative.

\subsection{Two-Stream Evidence Representation}

Using only the last-token stream reduces AUROC from 92.36 to 90.48, while using only the changed-token stream reduces it to 89.93. The two signals are therefore complementary rather than redundant.

\subsection{Step-Layer Embeddings}

Removing step embeddings reduces AUROC to 90.93, and removing layer embeddings reduces it to 91.43. This shows that selection depends not only on hidden content but also on where that content occurs in the denoising trajectory; the overall comparison is visualized in main-text Figure~\ref{fig:ablation_summary}.

\subsection{Decision-Logit Scoring}

The hard binary-decision variant reaches 86.53 F1, 82.69 balanced accuracy, 77.95 specificity, and 84.94 decision accuracy, all below the default decision-logit readout. More importantly, the logit readout yields a continuous score and therefore supports AUROC and AUPRC.

\section{Complete Result Tables}
\label{app:complete_results}

This appendix section reports the complete result tables corresponding to the compact summaries shown in the main paper. Specifically, we provide the full overall hallucination detection results, thresholded classification results on HotpotQA and TriviaQA, the full structured verification results, and the complete metric breakdowns for the primary Dream-7B-Instruct + TriviaQA ablation suite. The targeted LLaDA-8B-Instruct + HotpotQA interventions are reported in full in Table~\ref{tab:second_setting_ablation}. These tables are included here to keep the main paper focused on the most important comparisons while preserving complete numerical coverage for reference.

\begin{table*}[t]
\centering
\small
\setlength{\tabcolsep}{4pt}
\resizebox{\textwidth}{!}{
\begin{tabular}{l|ccc|ccc}
\toprule
& \multicolumn{3}{c|}{Dream-7B-Instruct} & \multicolumn{3}{c}{LLaDA-8B-Instruct} \\
\textbf{Method} & \textbf{HotpotQA} & \textbf{NQOpenLike} & \textbf{TriviaQA} & \textbf{HotpotQA} & \textbf{NQOpenLike} & \textbf{TriviaQA} \\
\midrule
Perplexity 
& 65.28 / 85.36 & 60.13 / 87.00 & 65.75 / 75.54
& 53.34 / 76.05 & 59.16 / 87.16 & 50.42 / 64.29 \\

LN-Entropy 
& 67.31 / 88.04 & 65.87 / 90.55 & 67.05 / 77.36
& 65.07 / 85.47 & 72.20 / 92.23 & 68.35 / 79.19 \\

Semantic Entropy 
& 81.84 / 93.10 & 72.17 / 91.47 & 81.92 / 86.92
& 78.00 / 90.04 & 81.24 / 94.57 & 80.98 / 86.79 \\

Lexical Similarity 
& 74.29 / 90.42 & 66.21 / 88.74 & 70.57 / 79.64
& 64.40 / 83.16 & 73.06 / 92.40 & 68.31 / 78.51 \\

EigenScore 
& 74.46 / 90.33 & 68.58 / 89.76 & 72.56 / 80.34
& 75.07 / 89.26 & 76.62 / 93.47 & 74.15 / 83.12 \\

CCS 
& 89.60 / 96.65 & 81.31 / 92.98 & 88.75 / 91.22
& 87.79 / 92.98 & 81.46 / 95.09 & 84.33 / 89.95 \\

TSV 
& 78.47 / 92.32 & 70.28 / 92.28 & 79.37 / 87.21
& 76.16 / 89.53 & 71.86 / 92.94 & 73.05 / 83.02 \\

TraceDet 
& 87.40 / 95.64 & 79.47 / 94.86 & 83.44 / 88.60
& 86.05 / 94.28 & 77.68 / 94.16 & 80.07 / 86.89 \\
\midrule

Text-only verifier
& 89.42 / 96.27
& 83.61 / 95.39
& 88.54 / 91.75
& 89.21 / 94.73
& 81.11 / 94.05
& 86.40 / 90.90 \\

\textbf{HIVE}
& \textbf{91.76 / 97.37} & \textbf{85.74 / 96.60} & \textbf{92.36 / 95.37}
& \textbf{90.94 / 96.48} & \textbf{85.71 / 96.95} & \textbf{90.66 / 93.98} \\
\bottomrule
\end{tabular}
}
\caption{\textbf{Full overall hallucination detection results.} Comparison against eight established baselines and the verifier-backbone-matched text-only verifier across two D-LLMs and three QA benchmarks. Each cell reports AUROC / AUPRC (\%, higher is better).}
\label{tab:appendix_full_main_results}
\end{table*}

\begin{table*}[t]
\centering
\small
\setlength{\tabcolsep}{4pt}
\resizebox{\textwidth}{!}{
\begin{tabular}{l|cc|cc}
\toprule
& \multicolumn{2}{c|}{Dream-7B-Instruct} & \multicolumn{2}{c}{LLaDA-8B-Instruct} \\
\textbf{Method} & \textbf{HotpotQA} & \textbf{TriviaQA} & \textbf{HotpotQA} & \textbf{TriviaQA} \\
\midrule
Perplexity
& 87.19 \; / \; 50.47 / 1.42
& 76.04 \; / \; 49.99 / 0.00
& 85.99 \; / \; 50.02 / 0.07
& 78.23 \; / \; 50.44 / 1.42 \\

LN-Entropy
& 87.27 \; / \; 50.01 / 0.02
& 76.04 \; / \; 49.99 / 0.00
& 85.96 \; / \; 50.42 / 1.21
& 78.20 \; / \; 52.66 / 8.43 \\

Semantic Entropy
& 88.11 \; / \; 64.63 / 35.87
& 80.30 \; / \; 70.49 / 54.87
& 85.99 \; / \; 50.00 / 0.00
& 82.23 \; / \; 74.03 / 64.43 \\

Lexical Similarity
& 87.37 \; / \; 50.25 / 0.55
& 76.35 \; / \; 51.80 / 4.97
& 86.49 \; / \; 58.07 / 20.13
& 79.76 \; / \; 60.53 / 28.28 \\

EigenScore
& 87.61 \; / \; 54.07 / 9.80
& 77.59 \; / \; 59.19 / 25.22
& 86.42 \; / \; 57.20 / 17.98
& 79.59 \; / \; 59.04 / 24.05 \\

CCS
& 90.53 \; / \; 73.72 / 53.54
& 83.94 \; / \; 79.18 / 74.36
& 90.30 \; / \; 74.13 / 53.45
& 83.58 \; / \; 73.51 / 58.78 \\

TSV
& 87.83 \; / \; 60.36 / 25.37
& 79.02 \; / \; 64.99 / 39.98
& 86.74 \; / \; 57.30 / 17.41
& 79.16 \; / \; 55.79 / 15.16 \\

TraceDet
& 89.78 \; / \; 76.48 / \textbf{62.68}
& 81.92 \; / \; 71.41 / 52.88
& 86.23 \; / \; 51.04 / 2.10
& 78.25 \; / \; 50.01 / 0.02 \\
\midrule
\textbf{HIVE}
& \textbf{91.47} \; / \; \textbf{77.77} / 61.96
& \textbf{87.76 \; / \; 83.63 / 78.53}
& \textbf{91.20 \; / \; 79.39 / 65.53}
& \textbf{88.07 \; / \; 83.27 / 78.41} \\
\bottomrule
\end{tabular}
}
\caption{\textbf{Thresholded classification results on HotpotQA and TriviaQA.} Each cell reports F1 / Balanced Accuracy / Specificity (\%, higher is better).}
\label{tab:appendix_full_thresholded_results}
\end{table*}

\begin{table*}[t]
\centering
\small
\setlength{\tabcolsep}{5pt}
\resizebox{\textwidth}{!}{
\begin{tabular}{l|ccc|ccc}
\toprule
& \multicolumn{3}{c|}{Dream-7B-Instruct} & \multicolumn{3}{c}{LLaDA-8B-Instruct} \\
\textbf{Metric} & \textbf{HotpotQA} & \textbf{NQOpenLike} & \textbf{TriviaQA} & \textbf{HotpotQA} & \textbf{NQOpenLike} & \textbf{TriviaQA} \\
\midrule
Valid-JSON Rate
& 100.00 & 99.83 & 100.00
& 100.00 & 99.43 & 100.00 \\

Decision Accuracy
& 86.46 & 83.88 & 85.29
& 86.49 & 83.18 & 84.55 \\

Hallucination-Type Accuracy
& 66.24 & 61.52 & 69.76
& 70.82 & 67.39 & 72.43 \\

Full-Schema Exact Match
& 58.87 & 47.34 & 65.86
& 53.60 & 31.11 & 60.49 \\
\bottomrule
\end{tabular}
}
\caption{\textbf{Full structured-output results of HIVE.}
Structured diagnostic output quality across all six model--benchmark settings (\%, higher is better). Valid-JSON Rate measures whether the verifier output is syntactically
well formed. Decision Accuracy measures correctness of the binary hallucination decision, and Hallucination-Type Accuracy measures agreement with the target type label.
Full-Schema Exact Match requires all evaluated output fields to match the reference target.}
\label{tab:appendix_full_structured_results}
\end{table*}

\begin{table*}[t]
\centering
\small
\setlength{\tabcolsep}{5pt}
\resizebox{\textwidth}{!}{
\begin{tabular}{lcccccc}
\toprule
\textbf{Variant} & \textbf{AUROC} & \textbf{AUPRC} & \textbf{F1} & \textbf{BAcc} & \textbf{Spec} & \textbf{DecAcc} \\
\midrule

\multicolumn{7}{l}{\textit{Ablation 1: Hidden-Evidence Prefix Conditioning}} \\
\quad HIVE (full) & \textbf{92.36} & \textbf{95.37} & \textbf{87.76} & \textbf{83.63} & \textbf{78.53} & \textbf{85.29} \\
\quad Plain verifier, no evidence prefix (retrained) & 88.54 & 91.75 & 84.25 & 78.28 & 74.62 & 81.25 \\
\quad Prefix-trained, no prefix at inference & 86.64 & 90.03 & 82.26 & 75.51 & 70.11 & 77.07 \\
\quad Shuffled evidence prefix & 88.57 & 90.75 & 84.28 & 78.17 & 74.30 & 81.18 \\

\midrule
\multicolumn{7}{l}{\textit{Ablation 2: Learned Step-Layer Evidence Selection}} \\
\quad HIVE (full learned selector) & \textbf{92.36} & \textbf{95.37} & \textbf{87.76} & \textbf{83.63} & \textbf{78.53} & \textbf{85.29} \\
\quad Final-step only & 89.17 & 93.39 & 85.82 & 79.27 & 73.25 & 81.94 \\
\quad Uniform over steps & 88.34 & 93.22 & 83.92 & 81.00 & 73.44 & 80.54 \\
\quad Random-$K$ evidence & 85.85 & 90.44 & 83.05 & 77.73 & 66.59 & 78.81 \\

\midrule
\multicolumn{7}{l}{\textit{Ablation 3: Two-Stream Evidence Representation}} \\
\quad Two-stream full $[u^{\mathrm{last}} ; u^{\mathrm{chg}}]$ & \textbf{92.36} & \textbf{95.37} & \textbf{87.76} & \textbf{83.63} & \textbf{78.53} & \textbf{85.29} \\
\quad Last-token only & 90.48 & 93.36 & 86.63 & 81.00 & 76.45 & 83.64 \\
\quad Changed-token only & 89.93 & 92.85 & 85.50 & 80.03 & 76.42 & 82.37 \\

\midrule
\multicolumn{7}{l}{\textit{Ablation 4: Step-Layer Embeddings}} \\
\quad Full (step + layer embeddings) & \textbf{92.36} & \textbf{95.37} & \textbf{87.76} & \textbf{83.63} & \textbf{78.53} & \textbf{85.29} \\
\quad w/o step embedding & 90.93 & 93.69 & 85.74 & 82.47 & 76.00 & 82.95 \\
\quad w/o layer embedding & 91.43 & 94.88 & 86.62 & 82.79 & 76.38 & 83.49 \\

\midrule
\multicolumn{7}{l}{\textit{Ablation 5: Decision-Logit Hallucination Scoring}} \\
\quad Decision-logit score & \textbf{92.36} & \textbf{95.37} & \textbf{87.76} & \textbf{83.63} & \textbf{78.53} & \textbf{85.29} \\
\quad Binary decision only & -- & -- & 86.53 & 82.69 & 77.95 & 84.94 \\

\bottomrule
\end{tabular}
}
\caption{\textbf{Full ablation results.} Complete metric breakdown for all ablation groups on Dream-7B-Instruct + TriviaQA. Metrics include AUROC, AUPRC, F1, Balanced Accuracy (BAcc), Specificity (Spec), and Decision Accuracy (DecAcc), all reported in \% (higher is better).}
\label{tab:appendix_full_ablation_results}
\end{table*}

\section{Asset Sources and Licenses}
\label{app:asset_licenses}

Table~\ref{tab:asset_licenses} summarizes the external models and datasets used in this work, together with their original sources and applicable licenses or usage terms. We cite the original creators of all such assets in the main paper and use them only for research and evaluation purposes within the scope of their official release terms.

\begin{table*}[t]
\centering
\footnotesize
\renewcommand{\arraystretch}{1.2}
\setlength{\tabcolsep}{4pt}
\begin{tabular}{p{2.3cm} p{1.4cm} p{3.0cm} p{2.8cm} p{5.1cm}}
\toprule
\textbf{Asset} & \textbf{Type} & \textbf{Source / Citation} & \textbf{License / Terms} & \textbf{Usage in this paper} \\
\midrule

Dream-v0-Instruct-7B
& Model
& \citet{ye2025dream7b}
& Apache-2.0
& Target D-LLM backbone used for trajectory extraction and evaluation. \\

LLaDA-8B-Instruct
& Model
& \citet{nie2025llada}
& MIT
& Target D-LLM backbone used for trajectory extraction and evaluation. \\

Qwen2.5-7B-Instruct
& Model
& \citet{qwen2025technical}
& Apache-2.0
& Automatic judge for supervision construction and verifier backbone for evidence-conditioned structured verification. \\

TriviaQA
& Dataset
& \citet{joshi2017triviaqa}
& Apache-2.0
& Evaluation benchmark for open-domain question answering. \\

HotpotQA
& Dataset
& \citet{yang2018hotpotqa}
& CC BY-SA 4.0
& Evaluation benchmark for multi-hop question answering. \\

Natural Questions
& Dataset
& \citet{kwiatkowski2019naturalquestions}
& Apache-2.0
& Source benchmark family underlying the NQ-style evaluation setting. \\

NQOpenLike
& Derived evaluation setting
& Derived from the Natural Questions benchmark family \citep{kwiatkowski2019naturalquestions}
& Subject to the license and terms of the underlying Natural Questions source data
& NQ-derived open-ended QA collection containing 8,758 question--answer examples. \\

\bottomrule
\end{tabular}
\caption{\textbf{External assets used in this paper.} We cite the original sources of all external models and datasets used in our experiments and summarize their licenses or usage terms where applicable. For the derived NQOpenLike setting, the relevant terms follow those of the underlying source data.}
\label{tab:asset_licenses}
\end{table*}

\section{Inference Profiling Details}
\label{app:inference_profiling}

We profile the deployment-time components of HIVE on Dream-7B-Instruct + TriviaQA using 100 held-out test examples, one NVIDIA H100 80GB GPU, and batch size 1. Hidden-state capture measures the incremental trajectory-capture work performed during the original D-LLM generation pass. Selector inference measures reconstruction of the two-stream step-layer features, selector scoring, and top-$K$ extraction. Evidence-conditioned verifier inference measures hidden-evidence prefix construction and structured verifier decoding.

\begin{table*}[t]
\centering
\small
\setlength{\tabcolsep}{6pt}
\renewcommand{\arraystretch}{1.05}
\begin{tabular}{@{}lrrrr@{}}
\toprule
\textbf{Pipeline} & \textbf{Mean latency} & \textbf{P95 latency} & \textbf{Throughput} & \textbf{Peak GPU memory} \\
& \textbf{(ms/ex.)} & \textbf{(ms/ex.)} & \textbf{(ex./s)} & \textbf{(GB)} \\
\midrule
Base D-LLM & 1,284 & 1,570 & 0.78 & 23.7 \\
Text-only pipeline & 2,465 & 2,882 & 0.41 & 42.6 \\
Full HIVE & 3,241 & 3,557 & 0.31 & 47.5 \\
\bottomrule
\end{tabular}
\caption{\textbf{Full end-to-end inference profile.} Results use Dream-7B-Instruct + TriviaQA, 100 held-out test examples, one NVIDIA H100 80GB GPU, batch size 1, and exclude model loading and warm-up. Throughput is the reciprocal of mean per-example latency.}
\label{tab:appendix_inference_efficiency_full}
\end{table*}

\begin{table*}[t]
\centering
\small
\setlength{\tabcolsep}{6pt}
\renewcommand{\arraystretch}{1.05}
\begin{tabular}{@{}lrrrr@{}}
\toprule
\textbf{Component} & \textbf{Mean latency} & \textbf{P95 latency} & \textbf{Throughput} & \textbf{Memory} \\
& \textbf{(ms/ex.)} & \textbf{(ms/ex.)} & \textbf{(ex./s)} & \textbf{(GB)} \\
\midrule
Hidden-state capture & 126 & 158 & -- & +3.8 \\
Selector inference & 8 & 10 & 125.0 & +0.1 \\
Evidence-conditioned verifier & 1,730 & 1,895 & 0.58 & 22.4 \\
\bottomrule
\end{tabular}
\caption{\textbf{Component-level inference profiling.} Hidden-state capture and selector memory are reported as incremental peak GPU memory relative to the corresponding pipeline stage, as indicated by the plus sign. Verifier memory is the standalone peak during evidence-conditioned verifier inference. Components were profiled independently and are therefore not expected to sum exactly to the end-to-end pipeline measurements in Appendix Table~\ref{tab:appendix_inference_efficiency_full}.}
\label{tab:inference_component_profile}
\end{table*}

Hidden-state capture increases Base D-LLM generation latency by 9.81\% but reuses the original generation pass and therefore does not require regenerating the answer. Selector inference is lightweight, contributing only 8 ms, or approximately 0.25\% of Full HIVE's end-to-end latency. The evidence-conditioned verifier is the largest independently profiled post-generation component, taking 1,730 ms on average. The remaining difference between independently measured component times and end-to-end latency reflects pipeline-level data movement, synchronization, and orchestration that are not isolated as separate components.

\section{Compute Resources}
\label{app:compute_resources}

All experiments were conducted on NVIDIA H100 80GB GPUs, with up to 10 GPUs used concurrently across runs. End-to-end and component-level deployment profiles are reported in Appendix~\ref{app:inference_profiling}. This section instead reports the offline preprocessing, training, evidence-export, and evaluation costs of constructing the reported systems. Table~\ref{tab:compute_resources} summarizes the approximate wall-clock time of the main pipeline stages; these values are reproducibility-oriented estimates and may vary across model--benchmark settings.

\begin{table}[!t]
\centering
\footnotesize
\setlength{\tabcolsep}{3pt}
\renewcommand{\arraystretch}{1.08}
\begin{tabular}{
    @{}
    p{0.42\columnwidth}
    p{0.23\columnwidth}
    p{0.27\columnwidth}
    @{}
}
\toprule
\textbf{Stage}
& \textbf{Hardware}
& \textbf{Wall-clock time} \\
\midrule
Trajectory feature construction
& 1$\times$H100 80GB
& 40--70 h \\

Selector training
& 1$\times$H100 80GB
& 5--20 min \\

Evidence export / dataset packing
& 1$\times$H100 80GB
& $\sim$1 min \\

Verifier training
& 1$\times$H100 80GB
& 20--90 h \\

Evaluation
& 1$\times$H100 80GB
& 10--30 h \\
\bottomrule
\end{tabular}

\caption{\textbf{Approximate compute usage by pipeline stage.}
Times are reported per model--benchmark setting.}
\label{tab:compute_resources}
\end{table}

For a single model--benchmark setting, the full HIVE pipeline therefore required approximately 70--190 hours of end-to-end wall-clock time. Under a single-GPU accounting view, this corresponds to roughly 70--190 H100-GPU-hours per setting. Since the main paper reports six model--benchmark settings, the main experiments required at least approximately 420--1140 GPU-hours in total, excluding pilot runs, failed runs, and routine debugging overhead.

\end{document}